\PassOptionsToPackage{table,dvipsnames}{xcolor}
\documentclass[10pt]{article}

\usepackage[accepted]{tmlr}

\usepackage{amsmath,amsfonts,bm}

\def\eqref#1{equation~\ref{#1}}

\def\1{\bm{1}}

\def\vb{{\bm{b}}}

\def\vu{{\bm{u}}}

\def\vx{{\bm{x}}}

\def\mA{{\bm{A}}}

\def\mI{{\bm{I}}}

\def\mM{{\bm{M}}}

\def\mU{{\bm{U}}}
\def\mV{{\bm{V}}}
\def\mW{{\bm{W}}}
\def\mX{{\bm{X}}}
\def\mY{{\bm{Y}}}

\def\mLambda{{\bm{\Lambda}}}
\def\mSigma{{\bm{\Sigma}}}

\DeclareMathAlphabet{\mathsfit}{\encodingdefault}{\sfdefault}{m}{sl}
\SetMathAlphabet{\mathsfit}{bold}{\encodingdefault}{\sfdefault}{bx}{n}
\newcommand{\tens}[1]{\bm{\mathsfit{#1}}}

\def\tM{{\tens{M}}}

\def\tW{{\tens{W}}}
\def\tX{{\tens{X}}}
\def\tY{{\tens{Y}}}
\def\tZ{{\tens{Z}}}

\usepackage[
  colorlinks=true,
  linkcolor=blue,
  citecolor=blue,
  urlcolor=blue
]{hyperref}
\usepackage{url}
\usepackage{booktabs}
\usepackage{amsfonts}
\usepackage{nicefrac}
\usepackage{microtype}
\usepackage{xcolor}
\usepackage{nicematrix}

\usepackage{graphicx}
\usepackage{subcaption}
\usepackage{placeins}
\usepackage{multirow}
\usepackage{comment}
\usepackage{float}
\usepackage{listings}
\usepackage{wrapfig}
\usepackage{enumitem}
\usepackage{pgffor}
\usepackage{etoolbox}

\usepackage{amsmath}
\usepackage{amssymb}
\usepackage{mathtools}
\usepackage{amsthm}
\usepackage{pifont}
\usepackage[capitalize,noabbrev,nameinlink]{cleveref}
\usepackage{thmtools}
\usepackage{thm-restate}
\usepackage[textsize=tiny]{todonotes}

\lstdefinestyle{mypython}{
  language=Python,
  basicstyle=\ttfamily\footnotesize,
  keywordstyle=\color{blue},
  commentstyle=\color{teal},
  stringstyle=\color{red!60!black},
  showstringspaces=false,
  breaklines=true,
  frame=single,
  columns=fullflexible,
  keepspaces=true,
  tabsize=4
}

\usepackage{algorithm}
\usepackage{algpseudocode}

\newcommand{\Reshape}{\mathsf{reshape}}
\newcommand{\Permute}{\mathsf{permute}}
\newcommand{\SVDTrunc}[2]{\mathcal{T}^{\mathrm{SVD}}_{#1}\bigl(#2\bigr)}
\newcommand{\AATrunc}[2]{\mathcal{T}^{\mathrm{AA}}_{#1}\bigl(#2\bigr)}
\newcommand{\Trunc}[3]{\mathcal{T}^{#3}_{#1}\bigl(#2\bigr)}

\newcommand{\ExprOut}{\overline{O}}
\newcommand{\ExprInp}{\overline{I}}
\newcommand{\ExprParam}{\overline{P}}

\newcommand{\id}{\mathrm{id}}

\theoremstyle{plain}
\newtheorem{theorem}{Theorem}[section]
\newtheorem{proposition}[theorem]{Proposition}
\newtheorem{lemma}[theorem]{Lemma}

\theoremstyle{definition}
\newtheorem{definition}[theorem]{Definition}
\newtheorem{example}[theorem]{Example}

\theoremstyle{remark}

\usepackage[table,dvipsnames]{xcolor}
\usepackage{tikz}
\usetikzlibrary{decorations.pathreplacing,arrows.meta}
\usepackage{nicematrix}
\colorlet{LightRoyalBlue}{RoyalBlue!4}

\newif\ifdeanon
\deanontrue

\ifdeanon
  \newcommand{\githuburl}{https://github.com/davidgonmar/BALF}
  \newcommand{\acknowledgements}{
    \subsubsection*{Acknowledgments}
    I thank Fernando Sancho Caparrini and Miguel Ángel Martínez del Amor for their valuable comments and suggestions, which helped improve this manuscript. I acknowledge access to computational resources provided by the University of Seville, including an A100 40GB GPU donated through the NVIDIA Hardware Grant, as well as an RTX 4090, and servers managed by the DeepKnowledge group. I am also grateful for the support of the University of Seville through the \emph{Programa de Becas de Iniciación a la Investigación (VIIPPIT-2025-II.1)}. Finally, I thank the anonymous reviewers for their constructive feedback.
  }
\else
  \newcommand{\githuburl}{https://anonymous.4open.science/r/BALF-2AFC}
  \newcommand{\acknowledgements}{}
\fi

\title{BALF: Budgeted Activation-Aware Low-Rank Factorization for Fine-Tuning-Free Model Compression}

\ifdeanon
\author{\name David Gonz\'alez-Mart\'inez\thanks{Most of this work was carried out while the author was at Universidad de Sevilla. The other listed affiliations are current.}
\email david.gonzalez-martinez@tue.ellis.eu \\
\addr Department of Computer Science and Artificial Intelligence, Universidad de Sevilla \\
Max Planck Institute for Intelligent Systems \\
ELLIS Institute T\"ubingen \\
T\"ubingen AI Center}
\else
\author{\name Anonymous Author(s) \email anonymous@example.com \\
\addr Anonymous Institution}
\fi

\def\month{11}
\def\year{2026}
\def\openreview{\url{https://openreview.net/forum?id=zvewqh2j7F}}

\ExplSyntaxOn
\cs_new:Npn \dg_rspace:n #1
  {
    \mathbb{R}^{\clist_use:nn { #1 } { \times }}
  }
\NewDocumentCommand{\Rspace}{m}
  {
    \dg_rspace:n { #1 }
  }
\ExplSyntaxOff

\begin{document}

\maketitle

\begin{abstract}
Activation-aware low-rank factorization techniques yield strong compression results, but they are generally confined to linear layers, and existing whitening-based theory typically makes an implicit full-rank assumption on layer activations. We introduce a layer representation framework that allows us to understand and extend low-rank factorization beyond linear layers, including standard and grouped convolutions. Within this framework, we provide a broader characterization of whitening-based activation-aware factorization that naturally covers rank-deficient activations, and show that the resulting projection is optimal in terms of layer output distortion. We then use output distortion as a proxy to allocate per-layer ranks under explicit FLOP or parameter count budgets, with negligible overhead via a Lagrangian relaxation of the allocation problem. Together, these components form BALF, an end-to-end, fine-tuning-free pipeline for model compression. Across CNNs and vision transformers on CIFAR-10 and ImageNet-1K, BALF generally achieves higher accuracy than SVD-based factorization baselines at matched FLOP or parameter count targets, and remains competitive with other fine-tuning-free compression techniques.
\end{abstract}

\section{Introduction}\label{sec:intro}
Deep learning models achieve state-of-the-art results across many domains \citep{LeCun2015}, yet their computational and memory demands limit practical deployment, motivating the need for model compression methods. Among these, low-rank factorization is a popular family of methods. Traditional techniques use the singular value decomposition (SVD) to minimize the discrepancy between original and compressed parameters \citep{Jaderberg2014SpeedingUC, hua-etal-2023-dynamic}. However, such techniques often rely on fine-tuning and/or costly search procedures.

Recently, the rise of large language models (LLMs) has motivated activation-aware factorization techniques, which seek to minimize the expected output distortion at each layer \citep{wang2025svdllm, wang2025svdllmv2optimizingsingular, NEURIPS2021_f56de5ef}, often reducing the need for fine-tuning. These techniques predominantly focus on fully connected layers, which are central to transformers \citep{NIPS2017_3f5ee243}, leaving other layer types and architectures less explored. Additionally, the whitening-based theory backing these techniques \citep{wang2025svdllm, wang2025svdllmv2optimizingsingular} generally assumes full-rank activations implicitly.

In this work, we propose a more general activation-aware factorization framework and introduce an efficient rank allocator that, given FLOP or parameter count budgets, determines the compression ratio of each layer with negligible compression overhead. Together, these form BALF, a principled fine-tuning-free compression pipeline for vision models. More concretely, our main contributions and findings are:
\begin{itemize}[noitemsep, topsep=0pt]
\item \textbf{Generalizing activation-aware factorization.} We introduce $(\ExprOut, \ExprInp, \ExprParam)$-expressible layers, a notion that enables us to extend matrix-based low-rank factorization to a broader class of layers, including standard and grouped convolutional layers \citep{Xie2016}. Under this framework, we generalize the whitening-based activation-aware theory of \citet{wang2025svdllm, wang2025svdllmv2optimizingsingular}. Our proposed approach covers a general class of whitening matrices and explicitly handles rank-deficient activations, avoiding ad hoc fixes such as adding a diagonal shift to maintain positive-definiteness \citep{wang2025svdllm}.
\item \textbf{Budget-aware allocation.} Using layer output distortion as a proxy for compression quality, we introduce an efficient rank allocator that determines the level of compression for each layer and approximately meets user-specified model-level FLOP or parameter count budgets. This is done via a Lagrangian relaxation of the combinatorial rank selection problem.
\item \textbf{Practicality.} BALF runs end-to-end efficiently on commodity hardware; for example, on an RTX 2080 Ti GPU, it takes less than four minutes to compress each model we tested. It also avoids expensive hyperparameter sweeps by directly targeting compression budgets.
\item \textbf{Empirical results.} Across vision models from ResNet-20 \citep{He_2016_CVPR} on CIFAR-10 \citep{Krizhevsky2009LearningML} to ResNeXt-101 \citep{Xie2016} and ViT-B/16 \citep{dosovitskiy2021an} on ImageNet-1K \citep{imagenet}, BALF generally surpasses SVD-based factorization baselines at matched FLOP or parameter count objectives and achieves competitive results against other fine-tuning-free compression techniques.
\end{itemize}

\section{Related Work}\label{sec:related_work}

Deep learning model compression has been extensively studied. The principal categories of compression techniques include quantization, which employs low-bit storage and operations \citep{Jacob_2018_CVPR, frantar2023optq, DBLP:journals/corr/abs-2103-13630}; pruning, which eliminates parameters or parameter groups from the model \citep{JMLR:v22:21-0366,frantar2022optimal}; and factorization, which decomposes model layers into smaller sub-layers to reduce memory usage and computational demands \citep{Jaderberg2014SpeedingUC, ben-noach-goldberg-2020-compressing}.

This paper focuses on factorization, of which many variants exist. Among these, the SVD \citep{golub2013matrix} stands out as a straightforward approach, as it can be used to obtain two low-rank factors that minimize the distortion in parameter space. It is directly applicable to fully connected layers, such as the linear projections commonly found in transformer architectures, which are implemented as matrix multiplications. In convolutional neural networks (CNNs), factorization is usually performed by reshaping parameter tensors into matrices and subsequently decomposing them \citep{DBLP:journals/corr/abs-2004-09031}. An alternative line of work employs other kinds of tensor decompositions \citep{10.5555/2968826.2968968, Yin_2021_CVPR, 10.1007/978-3-030-58526-6_31, liebenwein2021compressing, Lebedev2014SpeedingupCN}.

Existing factorization approaches often rely on fine-tuning and/or compute-intensive search \citep{learnc, yu2022svdnascouplinglowrankapproximation}. The difficulty of running such methods on common hardware motivates the need for lighter alternatives. For example, several pruning methods explicitly aim to be efficient in this regard \citep{narshana2023dfpc, murti2023tvsprune, chen2024going, zhang2024dense, wang2025forget}, whereas fine-tuning-free factorization methods are less common. One example is SVD-NAS \citep{yu2022svdnascouplinglowrankapproximation}, which still relies on an hours-long search procedure.

Recent factorization research, motivated by the computational demands of LLMs, leverages intermediate model features to identify additional redundancies suitable for decomposition \citep{NEURIPS2021_f56de5ef, wang2025svdllm, wang2025svdllmv2optimizingsingular, qinsi2025dobisvd, yuan2025asvdactivationawaresingularvalue}, substantially surpassing the compression--accuracy tradeoff of standard SVD-based approaches. These methods, which we refer to as ``activation-aware'', use a calibration dataset to characterize the distribution of activation tensors, and then use that information to guide parameter decompositions, often reducing the need for post-compression fine-tuning. However, they primarily focus on decomposing fully connected layers in language models, often overlooking other layers like convolutions, and settings such as image classification.\footnote{\citet{10.1109/TPAMI.2015.2502579} explore a similar data-aware approach in the image classification domain, but they also take nonlinearities into account and target older architectures.}

Several formulations of activation-aware factorization have been proposed. SVD-LLM \citep{wang2025svdllm, wang2025svdllmv2optimizingsingular} uses a whitening-based formulation, but its theory implicitly assumes full-rank activations, while others, such as DRONE \citep{NEURIPS2021_f56de5ef}, allow for rank-deficient activations without explicitly relying on whitening. Further extensions additionally account for compression shifts in subsequent layers \citep{sinha2026aasvdanchoredadaptive}. Similar problems are also studied in the field of reduced-rank regression \citep{reinsel2022multivariate}, although in a general setting rather than in the context of model compression. For convolutional layers, \citet{kalle2025distributionaware} introduces iterative activation-aware schemes based on tensor Tucker and CP decompositions. Our formulation builds on the whitening-based view of SVD-LLM. We extend it to a broader class of whitening matrices that explicitly allow for rank-deficient activations while providing a closed-form, matrix-based projection that is optimal in terms of layer output distortion and supports a general class of layers.

Regarding rank selection, which concerns the problem of selecting how much to compress each model layer (sometimes under a desired compute budget for the compressed model), approaches range from uniform rank selection \citep{wang2025svdllm,NEURIPS2021_f56de5ef} to more elaborate criteria \citep{abbasi2026zerosumsvdbalancing,wang2025svdllmv2optimizingsingular,10.1109/TPAMI.2015.2502579}, including training-based search \citep{qinsi2025dobisvd,learnc}, among many others. Here, we show that a simple approach based on minimizing layer output distortions can work well, while directly targeting user-specified budgets and remaining efficient via a Lagrangian relaxation of the combinatorial problem. Concurrent with the development of this work, \citet{ha2026afora} proposes a rank selection approach for activation-aware model compression that is very similar to ours, but focuses on LLMs.

\section{Preliminaries and Notation}\label{sec:preliminaries}

\paragraph{The singular value decomposition.} Given a matrix $\mA \in \Rspace{M, N}$, we denote its SVD by $\mA = \mU \mSigma \mV^\top$, where $\mU \in \Rspace{M, M}$ and $\mV \in \Rspace{N, N}$ are orthogonal matrices, and $\mSigma \in \Rspace{M,N}$ is a diagonal matrix containing the singular values $\bm{\sigma}$. We assume the singular values are sorted in decreasing order. We also use the $P$-truncated SVD, which is defined by retaining only the $P$ largest singular values and their associated singular vectors: $\SVDTrunc{P}{\mA} = \mU_{:,:P} \mSigma_{:P,:P} (\mV_{:,:P})^\top$. By the Eckart--Young--Mirsky (EYM) theorem, $\SVDTrunc{P}{\mA}$ is a best rank-$P$ approximation of $\mA$ in terms of the Frobenius norm \citep{golub2013matrix}.

\paragraph{Other notation and conventions.} We refer to general tensors as $\tX$. When a tensor is known to be a matrix or a vector, we may use $\mX$ and $\vx$, respectively. For matrices, we denote the Moore--Penrose inverse (also termed the pseudoinverse) of $\mM$ by $\mM^+$. For the identity function, we use $\mathrm{id}$. $\| \cdot \|_F$ refers to the tensor Frobenius norm, i.e., $\| \cdot \|_F = \|\mathsf{flatten}(\cdot)\|_2$.

Deep learning frameworks such as PyTorch \citep{Ansel_PyTorch_2_Faster_2024} usually implement ``reshape'' and ``permute'' operators, denoted in this work by $\Reshape\left(\tX; (s_1,\dots, s_n)\right)$ and $\Permute\left(\tX; (\pi_1,\dots,\pi_n)\right)$, respectively, where $(s_1,\dots,s_n)$ is a target shape and $(\pi_1,\dots,\pi_n)$ is some (in this work, zero-indexed) reordering of the dimensions of the original tensor. Moreover, when discussing convolutional layers, we use $H_x, W_x, C_x$ to denote the height, width, and channels of the input when $x=i$ and output when $x=o$; and $H_k, W_k$ to denote the spatial size of the kernel. In this context, $G$ denotes the number of groups in the convolutional layer \citep{Xie2016} ($G=1$ corresponds to an ungrouped convolution). We denote the convolution operator by $\tX*\tW$. 

When referring to neural network layers, unless otherwise specified, we drop layer identifiers and write the computation carried out by a layer as $f(\tX; \tW)$, where $\tX$ denotes the layer input and $\tW$ denotes the layer parameters. In the case of the first layer, $\tX$ denotes the input to the network (e.g., a multi-channel 2D signal). In general, when discussing factorization techniques, we ignore bias terms because they are unaffected by the factorization schemes considered in this work; only the primary operations (e.g., convolutions or matrix multiplications) are modified.

\section{Our Framework}\label{sec:method}

We present the different components of BALF in steps. We first introduce a framework that allows us to think about and extend low-rank factorization beyond linear layers. Within it, we then derive an optimal activation-aware projection. Finally, we show how we use the resulting output distortions to allocate per-layer ranks under a global complexity budget.

\subsection{Generalizing Low-Rank Factorization}

\paragraph{Low-rank projections and decompositions.} A matrix of rank $P$ or less can be expressed as the product of a tall matrix with $P$ columns and a wide matrix with $P$ rows. These two factors can be obtained exactly from the SVD. Concretely, if $\mA \in \Rspace{M, N}$ has rank $P$ or less, only its $P$ largest singular values can be nonzero, so $\mA = \SVDTrunc{P}{\mA} = \mA_0 \mA_1$, with $\mA_0 = \mU_{:,:P}\mSigma_{:P,:P}^{1/2} \in \Rspace{M, P}$ and $\mA_1 = \mSigma_{:P,:P}^{1/2}(\mV_{:,:P})^\top \in \Rspace{P, N}$. Hence, obtaining two such factors of a general matrix (not necessarily of rank $P$ or less) can be understood as first applying a low-rank projection that gives a matrix of rank at most $P$ (here denoted by $\Trunc{P}{\cdot}{}$) and then computing the decomposition through an SVD (any other way of extracting two factors exactly from a low-rank matrix also works). What determines the properties of the entire factorization scheme (encompassing both steps) is the projection $\Trunc{P}{\cdot}{}$.

This can be applied to compress model layers. We first focus on fully connected layers, which compute $\mX \mW$ for a given input $\mX \in \Rspace{B, I}$, where $B$ is the batch size, and weights $\mW \in \Rspace{I, O}$. If we fix $P$ and approximate $\mX \mW \approx \mX \Trunc{P}{\mW}{}$, we can then factorize the approximate layer exactly into $(\mX \mW_0) \mW_1$, with $\mW_0 \in \Rspace{I, P}$ and $\mW_1 \in \Rspace{P, O}$ obtained from an SVD of $\Trunc{P}{\mW}{}$, as explained earlier. For sufficiently small $P$, the number of parameters and FLOPs will be less than those of the original layer (\cref{sec:thorough_decomposition}; also note that the order of operations for the decomposed layer is important). A choice for the projection $\Trunc{P}{\cdot}{}$, among many, is the truncated SVD $\SVDTrunc{P}{\cdot}$ (see \cref{sec:preliminaries}), which minimizes $\| \mW - \SVDTrunc{P}{\mW}\|_F = \| \mW - \mW_0\mW_1 \|_F$, i.e., the distortion in parameter space. Next, we discuss how to extend the notion of factorization to other layers.

\paragraph{Expressing layers as matrix multiplications.} Low-rank factorization applies naturally to linear layers by decomposing their weight matrices. The main idea of our framework is to extend the notion of factorization and projection to $(\ExprOut, \ExprInp, \ExprParam)$-expressible layers, as defined below, by exploiting their equivalence to batched matrix multiplications.

\begin{definition}\label{def:oip_expressible}A layer $f$ is $(\ExprOut, \ExprInp, \ExprParam)$-expressible if it can be expressed as $f(\tX; \tW) = \ExprOut(\ExprInp(\tX)\ExprParam(\tW))$, where $\ExprOut$ and $\ExprParam$ are compositions of reshape and permute operators, and $\ExprInp$ is a linear operator. $\ExprOut$ takes a batch of matrices in the form of a third-order tensor and outputs a tensor (possibly a matrix), whereas $\ExprParam$ and $\ExprInp$ both take a tensor (possibly a matrix) and output a batch of matrices in the form of a third-order tensor of compatible shapes. In particular, $\ExprOut : \Rspace{G, N, O} \to \cdot$, $\ExprInp : \cdot \to \Rspace{G, N, I}$, and $\ExprParam: \cdot \to \Rspace{G, I, O}$.
\end{definition}

Here, $G$ stands for the number of ``groups''.\footnote{We use ``groups'' in analogy with grouped convolutions \citep{Xie2016}.} In the special case where $G=1$, we may simply refer to a ``matrix'' instead of ``a batch of a single matrix''. Note that in the preceding definition, matrix multiplication along the group dimension is batched, i.e., $G$ independent matrix multiplications are carried out, one per group. In what follows, layers themselves are assumed to operate in batched mode, with $B$ denoting the batch size (the following examples illustrate this). We refer to $N$ as the outer dimension. As a first example, a linear layer can be trivially expressed this way.

\begin{example}\label{example:oip_expr_linear}Fully connected layers are $(\id, \id, \id)$-expressible.
\end{example}

In some architectures (e.g., transformers), linear layers receive inputs with multiple leading independent dimensions (for example, a tensor of shape $(B, L, D)$). In those cases, we flatten the leading dimensions, reshaping the input to $(B L, D)$, treating $BL$ as the batch size, and restoring the original shape afterward. We now turn our attention to convolutional layers, which can also be expressed as in \cref{def:oip_expressible}. In particular, we consider standard (ungrouped) convolutions first.

\begin{example}\label{example:oip_expr_conv}
Ungrouped convolutional layers are $(\ExprOut,\ExprInp,\ExprParam)$-expressible, with
\begin{align*}
  \ExprOut(\tY) & = \mathsf{permute}(
    \mathsf{reshape}(\tY; (B, H_o, W_o, C_o)); (0, 3, 1, 2)) \in \Rspace{B, C_o, H_o, W_o}, \\
  \ExprInp(\tX) & = \mathsf{reshape}(
    \mathsf{im2col}(\tX); (B H_o W_o, C_i H_k W_k)) \in \Rspace{B H_o W_o, C_i H_k W_k}, \\
  \ExprParam(\tW) & = \mathsf{reshape}(\tW;(C_o, C_i H_k W_k))^\top \in \Rspace{C_i H_k W_k, C_o}.
\end{align*}
\end{example}

Here, $\mathsf{im2col}:\Rspace{B, C_i, H_i, W_i} \to \Rspace{B, H_o W_o, C_i H_k W_k}$ maps each 2D signal with $C_i$ channels in the batch into a matrix with each row corresponding to a convolutional window (see \citet{chellapilla:inria-00112631}). The PyTorch equivalent is called $\mathsf{unfold}$.\footnote{See \url{https://docs.pytorch.org/docs/stable/generated/torch.nn.Unfold.html}. GPTQ \citep{frantar2023optq} employs a similar formulation to handle convolutional layers in their code (\url{https://github.com/IST-DASLab/gptq/blob/main/gptq.py}), although they do not cover grouped layers. Similarly, \citet{10.1109/TPAMI.2015.2502579} roughly describe a similar idea.} It encapsulates the convolution hyperparameters---kernel size, stride, padding, and dilation---but not the number of groups. For brevity, we omit the encapsulated hyperparameters from the notation.

More generally, a grouped convolution can be viewed as $G$ separate, parallel convolutions. Each group takes a disjoint slice of \(C_i/G\) input channels and produces $C_o/G$ output channels. Concatenating the $G$ outputs recovers the full $C_o$ channels \citep{NIPS2012_c399862d, Xie2016} (note that $G=1$ corresponds to an ungrouped convolution). As shown below, grouped convolutions can also be expressed within our framework.

\begin{example}\label{example:oip_expr_conv_group}
Grouped convolutional layers are $(\ExprOut,\ExprInp,\ExprParam)$-expressible, with
  \begin{align*}
      \ExprOut(\tY) & = \mathsf{permute}(\mathsf{reshape}(\mathsf{permute}(\tY; (0, 2, 1));(C_o, B, H_o, W_o)); (1, 0, 2, 3)) \in \Rspace{B, C_o, H_o, W_o}, \\
      \ExprInp(\tX) & = \mathsf{permute}(\mathsf{reshape}(\mathsf{im2col}(\tX); (B H_o W_o, G, (C_i / G) H_k W_k)); (1, 0, 2)) \in \Rspace{G, B H_o W_o, (C_i / G) H_k W_k}, \\
      \ExprParam(\tW) & = \mathsf{permute}(\mathsf{reshape}(\tW;\,(G, C_o / G, (C_i / G) H_k W_k)); (0, 2, 1)) \in \Rspace{G, (C_i / G) H_k W_k, C_o / G}.
  \end{align*}
\end{example} 

\input{img/conv_as_matmul}

Additionally, \cref{example:oip_expr_conv} and \cref{example:oip_expr_conv_group} have trivial extensions to higher-dimensional convolutions, which can be obtained by replacing $\mathsf{im2col}$ with its higher-dimensional counterparts. 

\paragraph{Layer-general projections and decompositions.} Our key idea is that if a layer can be expressed as a matrix multiplication of transformed inputs and weights, we can define a corresponding low-rank layer by low-rank projecting the reshaped weights. Concretely, for an $(\ExprOut, \ExprInp, \ExprParam)$-expressible layer, we first low-rank project each matrix $\ExprParam(\tW)_g$ corresponding to a group, and then decompose the result as described above into $(\mW_0)_g \in \Rspace{I, P}$ and $(\mW_1)_g \in \Rspace{P, O}$, i.e., $\Trunc{P}{\ExprParam(\tW)_g}{} = (\mW_0)_g(\mW_1)_g$. For a small enough truncation rank $P$, storing these factors separately is always cheaper than storing $\tW$ (note that $\tW$ has the same number of elements as all $\ExprParam(\tW)_g$ combined). When the layer is needed in its original form, the (matrix) factors can be multiplied back together and mapped back with $\ExprParam{}^{-1}$ (denoting the inverse of $\ExprParam$), yielding $\Trunc{P}{\tW}{} \equiv \ExprParam{}^{-1}\left(\Trunc{P}{\ExprParam(\tW)}{}\right)$, where, abusing notation, $\Trunc{P}{\ExprParam(\tW)}{}$ denotes projecting each group's matrix individually and concatenating the results. We use this convention for all projection operators. We assume that the per-group rank is the same across groups. This constraint has practical reasons, which are discussed in \cref{subsec:grouped_conv_factors}.

Although we can save storage, running the decomposed layers naively in their matrix multiplication form can be expensive (as they require materializing $\ExprInp(\tX)$). The decomposition can also be compute-wise useful, but only if the layer admits an efficient decomposition. This is the case, for instance, for convolutional layers. In particular, it can be verified (see \cref{fig:conv_as_matmul} for a visual explanation) that $\tX * \tW \approx \tX * \Trunc{P}{\tW}{} = (\tX * \tW_0) * \tW_1$ for some $\tW_0 \in \Rspace{P, C_i, H_k, W_k}, \tW_1 \in \Rspace{C_o, P, 1, 1}$. Here, the underlying factorization is $\ExprParam(\tW)\approx \Trunc{P}{\ExprParam(\tW)}{} = \mW_0\mW_1$ with $\mW_0\in\Rspace{C_iH_kW_k,P}$ and $\mW_1\in\Rspace{P,C_o}$. These two factors can then be applied as two convolutions, with $\tW_0$ and $\tW_1$ being reshapes of $\mW_0^\top$ and $\mW_1^\top$, respectively. With sufficiently small $P$, this results in compute savings as well. All the instances explored here (linear and convolutional layers) have this property. \Cref{sec:thorough_decomposition} compares FLOPs and storage numbers before and after factorization for different layers.

\paragraph{Objective-based low-rank projections.} As discussed at the start of this section, what determines the properties of the factorization scheme is the projection operator $\Trunc{P}{\cdot}{}$, for which many options exist. When the projection is meant to minimize an objective, a general class of operators can be defined as 
\begin{equation}\label{eq:general_low_rank_proj}
\Trunc{P}{\tW}{D}
= \operatorname{arg}\,\underset{\operatorname{rank}(\ExprParam (\widehat{\tW})) \leq P}{\min}\;
   D(\tW, \widehat{\tW}),
\end{equation}
where $D$ is some penalty function, $\widehat{\tW}$ is the approximate tensor, and, for a batch of matrices, $\operatorname{rank}(\cdot) \leq P$ means that each matrix in the batch has rank at most $P$. We use equality to denote an arbitrary choice from the argmin when it is non-unique. The truncated SVD $\SVDTrunc{P}{\cdot}$ then corresponds to the special case where $D$ is the Frobenius norm (by the Eckart--Young--Mirsky (EYM) theorem, SVD solves the minimization problem \citep{golub2013matrix}), i.e., it penalizes distance in parameter space, but this might not be ideal. We next study a scheme that minimizes distortion in output space instead of parameter space.

\subsection{Activation-Aware Low-Rank Projections}

\paragraph{Motivation.} A more natural proxy for task performance is the (expected) distortion in each layer's outputs. We assume that we have a calibration dataset (any unlabeled set of i.i.d.~samples, for instance, the training dataset or a subset of it), which will be used to gather per-layer activations $(\tX^{(i)})_{i=1}^B$ (we omit the layer index for brevity), which we then use to estimate the distortion. More concretely, the (empirical) average output distortion for a layer $f$ and an approximation $\widehat{\tW}$ of its weights is defined as
\begin{equation}\label{eq:actdis_empirical}
\begin{aligned}
\ell^{\mathrm{out}}(\tW, \widehat{\tW}) =  \frac{1}{B} \sum_{i = 1}^B\|f(\tX^{(i)}; \tW) - f(\tX^{(i)}; \widehat{\tW})\|^2_F = \frac{1}{B} \|f(\tX; \tW) - f(\tX; \widehat{\tW})\|^2_F.
\end{aligned}
\end{equation}
The last equality holds when forming a ``batch'' with the list $(\tX^{(i)})_{i=1}^B$. From now on, our analysis is performed in terms of the intermediate activations $\tX$ obtained with the calibration dataset, with the expectation that the results carry over to the test distribution. We now aim to find a low-rank projection operator $\mathcal{T}_P$ that minimizes $\ell^{\mathrm{out}}(\tW, \Trunc{P}{\tW}{})$, i.e., an operator as in \cref{eq:general_low_rank_proj} with $D$ set to $\ell^{\mathrm{out}}$. Since $\ExprOut$ only rearranges entries (and thus preserves the Frobenius norm), the preceding formula can be written as
\[
\ell^{\mathrm{out}}(\tW, \widehat{\tW}) = \frac{1}{B}\big\|\ExprInp(\tX)\big(\ExprParam(\tW) - \ExprParam(\widehat{\tW})\big)\big\|^2_F .
\]

\paragraph{Which projection is best?} Without loss of generality, fix a group $g$. Clearly, $\operatorname{rank}\left(\ExprInp (\tX) \Trunc{P}{\ExprParam(\tW)}{} \right) \leq P$. Moreover, by the EYM theorem, the best rank-$P$ approximation of the output of $f$ in terms of the Frobenius norm is $\SVDTrunc{P}{\ExprInp (\tX ) \ExprParam (\tW)}$. Hence, a reasonable notion of ``ideal'' is a scheme that factorizes the parameters in a way that is equivalent to factorizing the outputs of the layer (it is clear that something better is impossible).
\begin{definition}\label{definition:optimality}
We call a scheme $\Trunc{P}{\cdot}{}$ optimal if its distortion equals that of directly low-rank truncating the outputs of a layer, i.e., an optimal scheme satisfies
\begin{align*}
\| \ExprInp(\tX) \ExprParam(\tW) - \SVDTrunc{P}{\ExprInp(\tX)\ExprParam(\tW)}\|_F = \| \ExprInp(\tX) \ExprParam(\tW) - \ExprInp(\tX) \Trunc{P}{\ExprParam(\tW)}{}\|_F.
\end{align*}
An alternative and equivalent definition is that the scheme satisfies \cref{eq:general_low_rank_proj} with the distance function set to the output distortion.
\end{definition}

\paragraph{Uncentered whitening.} In order to discuss activation-aware factorization, we first introduce our notion of whitening. Data whitening \citep{Kessy02102018} (also termed data sphering) is a widely used technique in machine learning and statistics. Here, we are interested in a special case called uncentered whitening. Different definitions exist, and they often assume full-rank data. Formally, we define an uncentered whitening matrix (UWM) $\mM \in \Rspace{I, I}$ for data $\mX \in \Rspace{N, I}$ as a matrix such that $\mM^\top \mX^\top \mX \mM = N \overline{\mI}_R$, where $\overline{\mI}_{R} \in \Rspace{I, I}$ is the identity on the first $R$ coordinates and zero elsewhere, and $R$ is the rank of $\mX$. We further restrict the definition by requiring that $\operatorname{row}(\mX) = \operatorname{range}(\mM)$, which in turn forces the last $I - R$ columns of $\mM$ to be zero and makes $\mX = \mX \mM \mM^+$. For a formal discussion, see \cref{sec:app_formal_uwm}.

Intuitively, plain SVD treats all weight (singular) directions equally, but weight errors along input directions with large activations distort the outputs much more than errors along directions with small activations. Whitening ``removes'' this imbalance. Fixing a single group and writing $\mX\mW = (\mX\mM)(\mM^+\mW)$, the whitened inputs $\mX\mM$ have uncorrelated features with the same ``energy'', thus every error in $\mM^+\mW$ distorts the outputs equally. This means that minimizing \cref{eq:actdis_empirical} amounts to minimizing the error of $\mM^+\mW$, which itself amounts to a truncated SVD on $\mM^+\mW$.

\paragraph{Our factorization scheme.} We now present our general version of whitening-based activation-aware projections under our framework. It satisfies \cref{definition:optimality}, allows for rank-deficient activations, and supports any UWM type as defined above. For any $(\ExprOut, \ExprInp, \ExprParam)$-expressible function $f$, let
\[
\AATrunc{P}{\tW}= \ExprParam{}^{-1}\left(\tM \SVDTrunc{P}{\tM^+\ExprParam\left(\tW\right)}\right),
\]
where $\tM$ is a batch of whitening matrices for $\ExprInp(\tX)$, one per group, and the pseudoinverse and matrix multiplications are performed independently across groups. For the layers discussed earlier, this again corresponds to two sequential sub-layers (e.g., two linear or convolutional layers). More formally, we can show that this scheme is optimal.

\begin{restatable}{proposition}{OptimalityProp}\label{prop:optimality} $\AATrunc{P} {\cdot}$ is optimal in the sense of \cref{definition:optimality}.
\end{restatable}
Moreover, the output distortion that follows from using our framework is readily computable from the artifacts obtained during compression.
\begin{restatable}{proposition}{GeneralActivationAwareDistortionProp}\label{prop:general_activation_aware_distortion}Assume that $f(\tX; \tW)$ is $(\ExprOut, \ExprInp, \ExprParam)$-expressible. Then, the output distortion (as defined in \cref{eq:actdis_empirical}) incurred when low-rank projecting the parameters with $\AATrunc{P}{\cdot}$  is $\frac{N}{B}\sum_{g=1}^{G}{\sum_{i=P+1}^U\bm{\sigma}_{g,i}^2}$, where $\bm{\sigma}$ denotes the batched singular values of $\tM^{+}\ExprParam(\tW)$ and $U$ is the total number of singular values per group.
\end{restatable}
Recall that $B$ is the batch size (number of calibration samples), and $N$ is the outer dimension (see the discussion below \cref{def:oip_expressible}).

\paragraph{Similarities and differences with close prior work. } SVD-LLM \citep{wang2025svdllm} introduced a formulation of activation-aware factorization for linear layers that targets the same objective (\cref{definition:optimality}). Because they rely on Cholesky-based whitening, their formulation implicitly assumes full-rank activations. In practice, they address rank-deficient cases with a small diagonal shift,\footnote{See \url{https://github.com/AIoT-MLSys-Lab/SVD-LLM}.} which is, in general, not optimal.\footnote{In practice, however, we expect no significant impact from this.} SVD-LLM V2 introduced a second scheme, but their results do not discuss the rank-deficient case (see \cref{sec:appendix_uwm}). In contrast, our formulation of whitening-based factorization is more general in the sense that we describe a general class of whitening matrices and allow for rank-deficient activations. Their main truncation loss results follow as corollaries. In particular, SVD-LLM and SVD-LLM V2 correspond to using Cholesky-based whitening and SVD-based whitening, respectively, both under the assumption of full-rank activations.\footnote{In our experiments (see \cref{sec:experiments}), Uniform-AA on linear layers attains the same projection error as SVD-LLM when activations are full rank, although we use our own exact formulation.}

\begin{wrapfigure}{r}{0.48\textwidth}
  \centering
  \vspace{-3.0em}
  \includegraphics[width=0.46\textwidth]{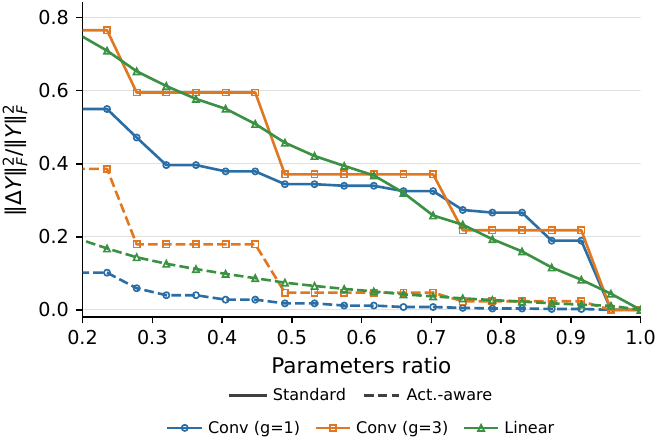}
  \caption{Normalized output distortion with $\AATrunc{P}{\cdot}$ and $\SVDTrunc{P}{\cdot}$ for different compression ratios, across different layer types. Activation-aware factorization consistently incurs significantly lower distortion.}
  \label{fig:rankratio_vs_output_frob}
  \vspace{-2.0em}
\end{wrapfigure}
\paragraph{A toy experiment.} We compare the output distortion of our activation-aware projection with that of standard SVD. In particular, we randomly initialize three layers of different types, truncate them to various parameter ratios, and measure the resulting output distortions on 1024 CIFAR-10 test images (using 1024 training images for calibration). As shown in \cref{fig:rankratio_vs_output_frob}, activation-aware factorization yields significantly lower distortion.

\paragraph{Bounding the output distortion of a model.} In \cref{prop:nn_bounds_prop_simple}, we show a network-level bound on the output distortion of sequential networks in terms of the individual distortions for each layer, $\ell_l^{\mathrm{out}} = \ell^{\mathrm{out}}(\tW^l, \widehat{\tW}{}^l)$, where $\widehat{\tW}{}^l = \AATrunc{P_l}{\tW^l}$. We denote the corresponding cumulatively distorted intermediate activations as $\widehat{\tX}{}^{l}$.

\begin{restatable}{proposition}{GeneralBoundsPropSimple}(Informal)\label{prop:nn_bounds_prop_simple}
Define a sequential $L$-layer network
\begin{align*}
\tX^{l}
= a_l\left(f_{l}(\tX^{l-1}; \tW^{l}) + \vb^l \right)
= a_l \left(\ExprOut_{l}\!\big(\ExprInp_{l}(\tX^{l-1})\ExprParam_{l}(\tW^{l})\big) + \vb^l\right),
\end{align*}
with $\tX^{0} = \tX$, where $a_l$ is an element-wise activation function with Lipschitz constant $A_l$. Moreover, let $B_l$ be the Lipschitz constant of $\ExprInp_{l}$ (recall that $\ExprInp_{l}$ is linear). We assume that all layers are ungrouped. Then,
\begin{equation*}
\begin{aligned}
&\frac{1}{\sqrt{B}}\bigl\| \tX^{L} - \widehat{\tX}{}^{L} \bigr\|_F \le
\sum_{l=1}^L \Bigl[ A_l\sqrt{\ell^{\mathrm{out}}_l}
\prod_{i=l+1}^L\|\ExprParam_i(\widehat{\tW}{}^{i})\|_2 A_iB_i \Bigr].
\end{aligned}
\end{equation*}
\end{restatable}

A more detailed statement and discussion, along with the proof and special cases, appear in \cref{sec:nn_distortion_bounds}. The tightness of the bound depends on the model's depth and its weights, and it can be loose in deep networks. Nevertheless, it intuitively supports keeping the individual layer distortions small, which we use for rank allocation below (see \cref{sec:rank_allocator_appendix} for a brief discussion on alternative choices for the allocation objective).

\subsection{Budgeted Rank Allocation}\label{subsec:selecting_rank_per_layer}

Apart from offering output distortion guarantees, our scheme makes each layer's output distortion computable without any additional step (see \cref{prop:general_activation_aware_distortion}). We now use this to efficiently choose the rank to which each layer is truncated.

\paragraph{The rank allocation problem. }The goal is to select the ranks to retain in each layer, $\{P_l\}_{l=1}^L$. The naive option is to choose a uniform compression ratio per layer \citep{wang2025svdllm}. A more reasonable option, used in the past for the traditional SVD scheme, is the energy-based singular value pruning criterion \citep{DBLP:journals/corr/abs-2004-09031,liebenwein2021compressing}. While more expensive approaches have been explored, such as training-based rank learning \citep{learnc, qinsi2025dobisvd} or NAS-based search \citep{yu2022svdnascouplinglowrankapproximation}, our main goal is to find a scalable and efficient method. We can define the analog of the energy metric in our setting as follows (it reads as the energy retained in layer $l$ when keeping up to rank $P_l$):
\begin{equation}\label{eq:energy}
E_l(P_l) = \frac{\sum_{g=1}^G\sum_{i=1}^{P_l} \bm{\sigma}_{g,i}^2}{\sum_{g=1}^G \sum_{i=1}^{U_l} \bm{\sigma}_{g,i}^2} = 1 - \frac{\|f_l(\tX^{l-1}; \tW^l) - f_l(\tX^{l-1}; \AATrunc{P_l}{\tW^l})\|^2_F}{\|f_l(\tX^{l-1}; \tW^l)\|^2_F},
\end{equation}
where $l$ denotes the layer index, $\bm{\sigma}$ the batched singular values of $(\tM^{l})^{+}\ExprParam(\tW^{l})$ (as in \cref{prop:general_activation_aware_distortion}), $U_l$ the total number of singular values (per group), and $P_l$ the number retained (per group). Energy-based selection then consists of picking the lowest rank such that a certain energy threshold (manually defined by the user) is retained. We propose instead to solve an optimization problem that maximizes the total retained energy globally:
\begin{gather*}
\underset{P_1, \dots, P_L}{\max}\; \sum_{l=1}^L E_l(P_l), \quad \text{subject to }\; \sum_{l=1}^L C_l(P_l) \leq C_{\max}
\;\text{and} \; P_l \in \{1,\dots,U_l\},
\end{gather*}
where $C_l(P_l)$ is some measure of layer complexity (in our case, FLOPs or number of parameters) when keeping up to rank $P_l$ on layer $l$, and $C_{\max}$ is the global model complexity budget. Our allocator supports FLOPs and the number of parameters as complexity targets. The optimization problem is an instance of the multiple-choice knapsack problem, and it can be formulated as a binary linear program. Unfortunately, it is known to be NP-hard (for a review, see \citet{math13071097}), which motivates the need for an alternative.

\paragraph{A Lagrangian relaxation.} We approach this limitation by using a Lagrangian relaxation of the problem to obtain approximate solutions. It has a time complexity of $O((T+ S + 1)\sum_l U_l)$, where $T$ is the number of iterations and $S$ depends on the spectral properties of the network. In practice, using 300 iterations suffices and takes less than 0.2 seconds for all models tested. We provide additional details in \cref{subsec:lagrangian}. We note that, concurrent with the development of this work, \citet{ha2026afora} proposed a very similar idea. Although they discuss it in the context of ``water filling'', the resulting algorithm is close to ours. 

\paragraph{Advantages. }Our selection approach has several desirable properties. In particular, (1) it has virtually zero additional overhead and does not query the model, unlike brute-force search methods, which can take hours (e.g., \citet{yu2022svdnascouplinglowrankapproximation}), and (2) it provides explicit control over a complexity budget. This reduces the need for manual hyperparameter sweeps to achieve a compression target.

\subsection{Putting It All Together: BALF}

Together, these pieces form BALF. An overview can be found in \cref{alg:balf_overview}. The routine $\textsc{NewLayer}$ generates a new layer (consisting of two sequential sub-layers) when given the factorized matrices. In practice, we do not select a manual budget $C_{\max}$, but rather a ratio (e.g., we might want to obtain a model with 50\% of its original FLOPs). Additionally, note that we retain the original layer if compressing it to the selected rank is not computationally advantageous.

\paragraph{Computing the whitening matrices in practice.}\label{subsec:whitening_matrices} Although our framework admits a general class of whitening matrices, in our experiments we compute them via eigendecompositions. Suppose we are given a group $g$. Let $\mX = \ExprInp(\tX)_g \in \Rspace{N, I}$. We can then compute the eigenvectors and eigenvalues of $\frac{1}{N}\mX^\top\mX$: $\mV$ and $\mLambda$, respectively, and set $\mM = \mV \mLambda^{+1/2}$ and $\mM^{+} = \mLambda^{1/2} \mV^\top$ (we abuse notation and use $\cdot^{+1/2}$ to denote the square root of the pseudoinverse). Here, the pseudoinverse can be computed by inverting individual nonzero (in practice, larger than a small threshold) entries, as $\mLambda$ is diagonal.

This procedure is applied (in vectorized form) independently for each group. Second moments can be accumulated incrementally over minibatches, making the method memory-efficient. Additional details are provided in \cref{sec:appendix_uwm}. One alternative is to compute whitening matrices sequentially, using the outputs of already-compressed preceding layers. Another way is to avoid full recomputation but update relevant elements using distorted inputs, similar to an older SVD-LLM version. In preliminary small-scale experiments, we found that recomputing whitening matrices sequentially offered no significant advantage. Similar to SVD-LLM, we therefore precompute the second moments, whitening matrices, and factors once; they are then cached and can be reused. This is significantly more efficient when generating multiple versions of the same model at different compression ratios (see \cref{sec:runtime_impl} for more details about the time cost of each step).

\begin{algorithm}[ht]
\caption{BALF overview. Group notation is omitted for simplicity.}
\label{alg:balf_overview}
\begingroup\small
\begin{algorithmic}[1]
\Require Pre-trained model $\{\mathcal{M}_l\}_{l=1}^{L}$, calibration data $\mathcal{D}$, budget $C_{\max}$, cost functions $\{{C_l}\}_{l=1}^{L}$
\Ensure Compressed model (compressed in place for simplicity)
\State $\{\frac{1}{N_l}\ExprInp_l(\tX^{l-1})^\top\ExprInp_l(\tX^{l-1})\}_{l=1}^{L} \gets \textproc{CollectActivations}(\{\mathcal{M}_l\}_{l=1}^{L},\mathcal{D})$ \Comment{stored in reshaped form}
\State $F \gets [\;]$
\For{$l=1,\dots,L$}
  \State $[\mM,\mM^{+}] \gets \textproc{BatchedUWM}(\frac{1}{N_l} \ExprInp_l(\tX^{l-1})^\top\ExprInp_l(\tX^{l-1}));\ \tW \gets \mathcal{M}_l.\operatorname{weights}$
  \State $[\mU,\mSigma,\mV^\top] \gets \textproc{BatchedSVD}(\mM^{+}\,\ExprParam_l(\tW))$
  \State $F[l] \gets [\mU,\mSigma,\mV^\top,\mM,\mM^{+}]$
\EndFor
\State $\{P_l\}_{l=1}^{L} \gets \textproc{AllocateRanks}(\textproc{GetSigmas}(F), C_{\max}, \{C_l\}_{l=1}^{L})$
\For{$l=1,\dots,L$}
  \State \algorithmicif\ {$C_l(P_l)\ge \textproc{OriginalCost}(l)$} \algorithmicthen\ \textbf{continue} \Comment{factorizing might not be worth it}
  \State $[\mU,\mSigma,\mV^\top,\mM,\mM^{+}] \gets F[l]$
  \State $\mW_0 \gets \mM\mU_{:,:P_l}\mSigma_{:P_l,:P_l}^{1/2}$;\quad $\mW_1 \gets \mSigma_{:P_l,:P_l}^{1/2}(\mV^\top)_{:P_l,:}$
  \State $\mathcal{M}_l \gets \textproc{NewLayer}(\mW_0,\mW_1,\mathcal{M}_l.\operatorname{bias},P_l)$ \Comment{constructs a new layer composed of two sequential layers}
\EndFor
\end{algorithmic}
\endgroup
\end{algorithm}

\paragraph{Handling different network structures.} Our method operates at the layer level: it modifies each layer's internal computation without changing inter-layer connections. As a result, it supports different architectures (e.g., residual connections) without special handling.

\section{Experiments}\label{sec:experiments}

This section presents our experimental results across a range of architectures, datasets, and scales. Our code is available at \url{\githuburl}. On CIFAR-10 \citep{Krizhevsky2009LearningML}, we evaluate ResNet-20 and ResNet-56, while on ImageNet-1K \citep{imagenet}, we consider ResNet-18 and ResNet-50 \citep{He_2016_CVPR}, ResNeXt-50 (32$\times$4d) and ResNeXt-101 (32$\times$8d) with grouped convolutions \citep{Xie2016}, ViT-B/16 \citep{dosovitskiy2021an}, DeiT-B/16 \citep{pmlr-v139-touvron21a}, and MobileNetV2 \citep{Sandler_2018_CVPR}. For all models, every factorizable layer is considered for factorization. Calibration for the main experiments uses 1024 images for CIFAR-10 and 8192 images for ImageNet-1K, both uniformly sampled from their respective training datasets. We report accuracy on the CIFAR-10 test set and the ImageNet-1K validation set. The precise settings and hyperparameters are provided in \cref{sec:experimental_details}.

\paragraph{Comparison with factorization baselines.}
We first compare BALF against different factorization baselines under approximately matched FLOP and parameter count budgets. We conduct a broad sweep over compression ratios to obtain complexity--accuracy tradeoff curves. We implement and compare different baselines under the same conditions:
\begin{itemize}[noitemsep, topsep=0pt]
    \item \textbf{Uniform and Uniform-AA.}
    Given some target ratio $\rho$, the same factorized-to-original parameter ratio is used for every layer. For layer $l$, the parameter and FLOP ratios obtained by retaining rank $P_l$ (the two ratios coincide) are $R_l(P_l)={P_l(I_l+O_l)}/{I_lO_l}$ (\cref{sec:thorough_decomposition}). We select $P_l$ so that $R_l(P_l) \approx \rho$ for every layer, up to rank discretization. Uniform and Uniform-AA use standard SVD and activation-aware SVD, respectively.
    \item \textbf{Energy and Energy-AA.}
    Given an energy threshold $\tau$, the rank of each layer is selected independently according to $P_l=\min\{P\in\{1,\ldots,U_l\}:E_l(P)\geq\tau\}$, with $E_l$ as in the first expression of \cref{eq:energy}.
    Energy uses the singular values of
    $\ExprParam(\tW^l)$ and standard SVD, while
 Energy-AA uses the singular values of
 $(\tM^{l})^{+}\ExprParam(\tW^l)$ and activation-aware SVD.
\end{itemize}
For BALF, which corresponds to activation-aware SVD with our rank allocator, we consider both FLOP objectives (BALF-F) and parameter count objectives (BALF-P). For all activation-aware methods, which depend on calibration data, we run three seeds (which affect the calibration sample) and report the mean accuracy. For BALF and Energy-AA, whose rank allocations also depend on the calibration data (different calibration samples can produce slightly different compression factors), each reported point denotes an average of accuracy and parameter/FLOP counts over the three calibration seeds. The variability induced by the calibration set is negligible, both in terms of compression and accuracy. Comparing against these baselines also allows us to isolate the behavior of different components of BALF. In particular, comparing Uniform and Energy with their activation-aware counterparts isolates the effect of the activation-aware projection. Similarly, comparing the activation-aware variants with BALF isolates the effect of our rank allocator.

\begin{figure*}[!ht]
  \centering
  \begin{subfigure}[t]{0.33\textwidth}
    \centering
    \includegraphics[width=\linewidth]{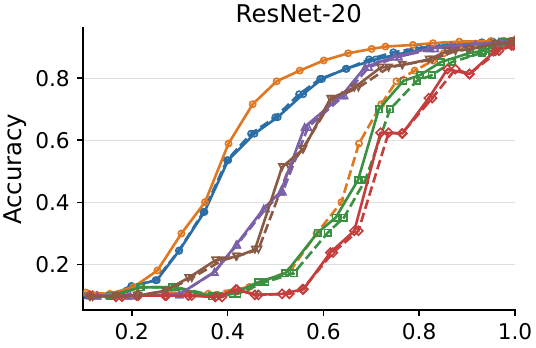}
  \end{subfigure}\hfill
  \begin{subfigure}[t]{0.33\textwidth}
    \centering
    \includegraphics[width=\linewidth]{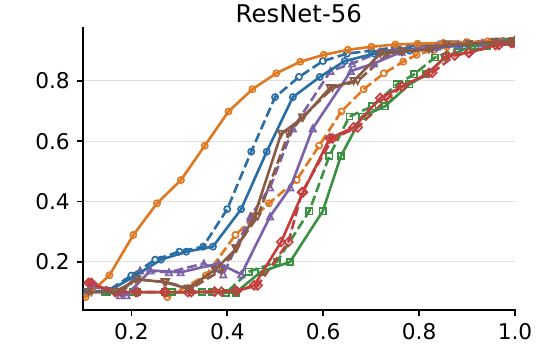}
  \end{subfigure}\hfill
  \begin{subfigure}[t]{0.33\textwidth}
    \centering
    \includegraphics[width=\linewidth]{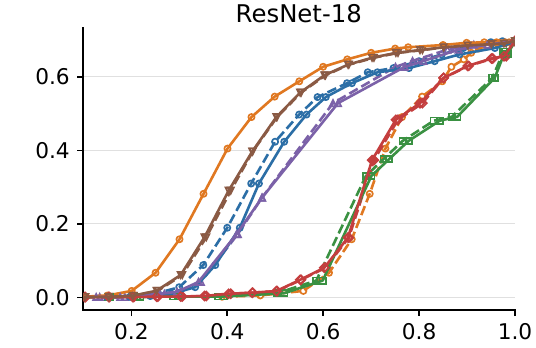}
  \end{subfigure}

  \vspace{0.5em}
  \begin{subfigure}[t]{0.33\textwidth}
    \centering
    \includegraphics[width=\linewidth]{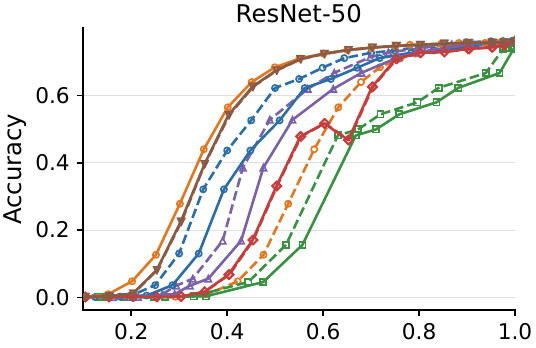}
  \end{subfigure}\hfill
  \begin{subfigure}[t]{0.33\textwidth}
    \centering
    \includegraphics[width=\linewidth]{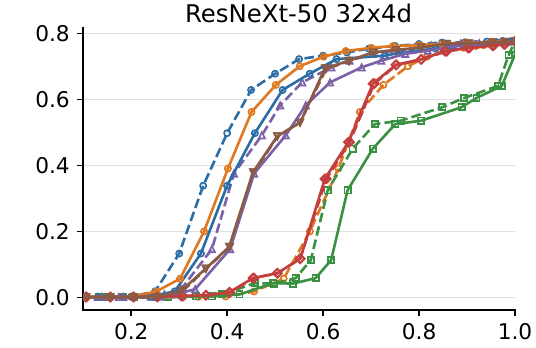}
  \end{subfigure}\hfill
  \begin{subfigure}[t]{0.33\textwidth}
    \centering
    \includegraphics[width=\linewidth]{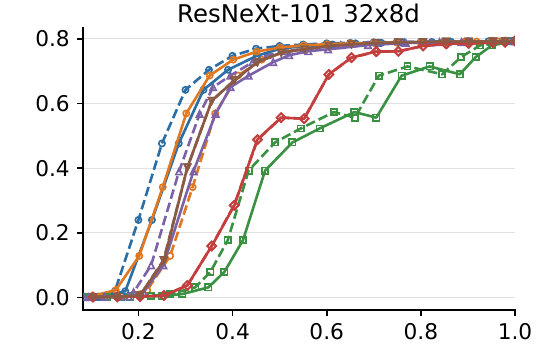}
  \end{subfigure}

  \vspace{0.5em}
  \begin{subfigure}[t]{0.33\textwidth}
    \centering
    \includegraphics[width=\linewidth]{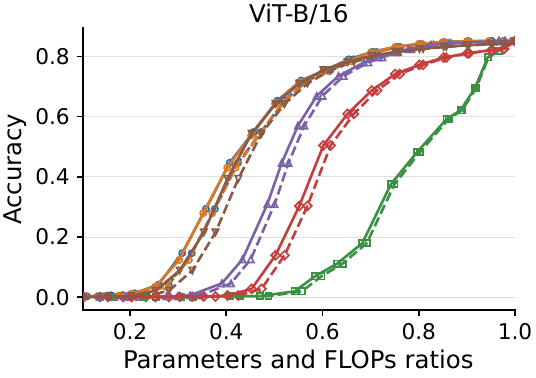}
  \end{subfigure}\hfill
  \begin{subfigure}[t]{0.33\textwidth}
    \centering
    \includegraphics[width=\linewidth]{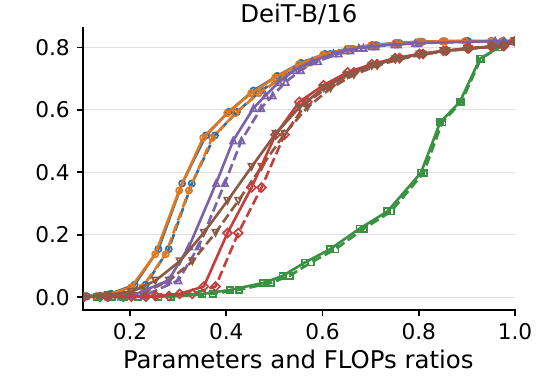}
  \end{subfigure}\hfill
  \begin{subfigure}[t]{0.33\textwidth}
    \centering
    \includegraphics[width=\linewidth]{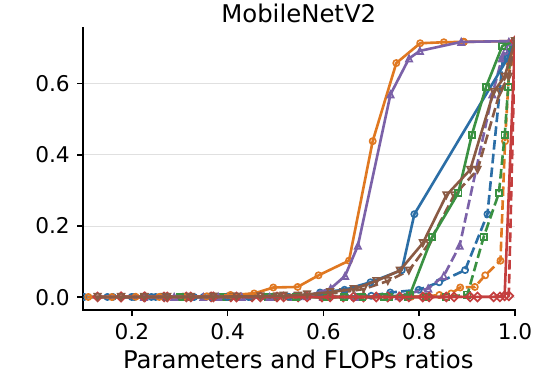}
  \end{subfigure}

  \vspace{0.2em}
  \includegraphics{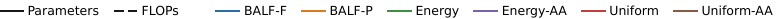}

\caption{Sweep over compression ratios for different factorization methods across models. Each panel shows parameter--accuracy and FLOP--accuracy curves for each method (denoted by solid and dashed lines, respectively). BALF-F and BALF-P denote BALF configured with FLOP and parameter count objectives, respectively.}
  \label{fig:posttraining-results}
\end{figure*}

{
\renewcommand{\arraystretch}{0.95}
\setlength{\tabcolsep}{2pt}

\begin{table*}[t]
\caption{Comparison of different methods at different compression configurations on ImageNet-1K models. BALF-P-$x$ and BALF-F-$x$ denote our method when set with parameter count and FLOP budgets, respectively, equal to a fraction $x$ of the original model's. ``$-$'' is used for unavailable results. Methods appearing multiple times correspond to different compression ratios. $\Delta$FLOPs\% and $\Delta$Params\% denote the percentage changes in FLOPs and parameter counts relative to the original model (a larger change means the model is compressed further), and $\Delta$Top-1 pp denotes the change in top-1 accuracy in percentage points. For BALF, we sample three seeds and report mean and standard deviation (random variability comes from the calibration subset).}
\label{tab:results_comparison}
\centering
\small

\begin{subtable}[t]{0.49\textwidth}
\caption{ResNet-18}
\centering
\begin{NiceTabular*}{\linewidth}{@{\extracolsep{\fill}}p{2.1cm}ccc@{}}
\CodeBefore
  \rectanglecolor{LightRoyalBlue}{12-1}{17-4}
\Body
\toprule
Method & $\Delta$FLOPs\% $\downarrow$ & $\Delta$Params\% $\downarrow$ & $\Delta$Top-1 pp $\uparrow$ \\
\midrule
T-ALS     & $-$ & $-28.57$ & $-10.90$ \\
T-ALS     & $-$ & $-33.33$ & $-17.60$ \\
T-ALS-S   & $-$ & $-28.57$ & $-3.00$ \\
T-ALS-S   & $-$ & $-43.50$ & $-6.50$ \\
CP-ALS    & $-$ & $-51.69$ & $-3.22$ \\
CP-ALS    & $-$ & $-60.47$ & $-8.03$ \\
CP-ALS-S  & $-$ & $-51.69$ & $-1.91$ \\
CP-ALS-S  & $-$ & $-69.23$ & $-6.70$ \\
SVD-NAS   & $-58.60$ & $-68.05$ & $-13.35$ \\
ALDS      & $-42.31$ & $-65.14$ & $-18.70$ \\
BALF-F-0.6 & $-41.18_{\pm 0.00}$ & $-39.45_{\pm 0.00}$ & $-15.25_{\pm 0.05}$ \\
BALF-P-0.6 & $-13.26_{\pm 0.00}$ & $-40.01_{\pm 0.00}$ & $-7.00_{\pm 0.02}$ \\
BALF-F-0.7 & $-30.71_{\pm 0.01}$ & $-28.68_{\pm 0.02}$ & $-8.50_{\pm 0.01}$ \\
BALF-P-0.7 & $-9.13_{\pm 0.01}$ & $-29.99_{\pm 0.01}$ & $-3.20_{\pm 0.03}$ \\
BALF-F-0.8 & $-20.01_{\pm 0.00}$ & $-16.74_{\pm 0.00}$ & $-5.49_{\pm 0.04}$ \\
BALF-P-0.8 & $-6.77_{\pm 0.01}$ & $-22.28_{\pm 0.02}$ & $-1.69_{\pm 0.02}$ \\
\bottomrule
\end{NiceTabular*}
\vspace{0.4em}
\end{subtable}
\hfill
\begin{subtable}[t]{0.49\textwidth}
\caption{ResNet-50}
\centering
\begin{NiceTabular*}{\linewidth}{@{\extracolsep{\fill}}p{2.1cm}ccc@{}}
\CodeBefore
  \rectanglecolor{LightRoyalBlue}{12-1}{17-4}
\Body
\toprule
Method & $\Delta$FLOPs\% $\downarrow$ & $\Delta$Params\% $\downarrow$ & $\Delta$Top-1 pp $\uparrow$ \\
\midrule
IFM       & $-$ & $-5.65$  & $-1.36$  \\
IFM       & $-$ & $-20.42$ & $-10.45$ \\
ITVSP     & $-$ & $-4.76$  & $-3.02$  \\
ITVSP     & $-$ & $-9.98$  & $-10.21$ \\
DFPC      & $-$ & $-5.66$  & $-5.78$  \\
DFPC      & $-$ & $-10.92$ & $-13.88$ \\
T-ALS     & $-$ & $-27.54$ & $-4.49$  \\
T-ALS     & $-$ & $-33.33$ & $-20.14$ \\
T-ALS-S   & $-$ & $-27.54$ & $-2.57$  \\
T-ALS-S   & $-$ & $-33.33$ & $-6.90$  \\
BALF-F-0.5 & $-50.03_{\pm 0.02}$ & $-43.84_{\pm 0.01}$ & $-13.96_{\pm 0.05}$ \\
BALF-P-0.5 & $-28.17_{\pm 0.01}$ & $-50.06_{\pm 0.01}$ & $-7.79_{\pm 0.05}$ \\
BALF-F-0.7 & $-30.01_{\pm 0.00}$ & $-21.66_{\pm 0.01}$ & $-3.50_{\pm 0.01}$ \\
BALF-P-0.7 & $-13.60_{\pm 0.00}$ & $-29.95_{\pm 0.00}$ & $-1.97_{\pm 0.03}$ \\
BALF-F-0.8 & $-20.03_{\pm 0.02}$ & $-9.73_{\pm 0.01}$ & $-1.69_{\pm 0.02}$ \\
BALF-P-0.8 & $-9.03_{\pm 0.01}$ & $-21.90_{\pm 0.01}$ & $-1.11_{\pm 0.01}$ \\

\bottomrule
\end{NiceTabular*}
\vspace{0.4em}
\end{subtable}

\par\vspace{0.4em}

\begin{subtable}[t]{0.49\textwidth}
\caption{DeiT-B/16}
\centering
\begin{NiceTabular*}{\linewidth}{@{\extracolsep{\fill}}p{2.1cm}ccc@{}}
\CodeBefore
  \rectanglecolor{LightRoyalBlue}{6-1}{11-4}
\Body
\toprule
Method & $\Delta$FLOPs\% $\downarrow$ & $\Delta$Params\% $\downarrow$ & $\Delta$Top-1 pp $\uparrow$ \\
\midrule
GTP       & $-13.07$ & $0.00$ & $0.00$ \\
GTP       & $-50.00$ & $0.00$ & $-3.50$ \\
PRACTISE  & $-16.6$ & $-$ & $-2.50$  \\
DC-ViT    & $-16.6$ & $-$ & $-0.54$ \\
BALF-F-0.6 & $-38.37_{\pm 0.00}$ & $-39.52_{\pm 0.00}$ & $-4.10_{\pm 0.03}$ \\
BALF-P-0.6 & $-38.71_{\pm 0.00}$ & $-39.87_{\pm 0.00}$ & $-4.31_{\pm 0.02}$ \\
BALF-F-0.7 & $-28.78_{\pm 0.00}$ & $-29.64_{\pm 0.00}$ & $-1.25_{\pm 0.05}$ \\
BALF-P-0.7 & $-29.04_{\pm 0.00}$ & $-29.91_{\pm 0.00}$ & $-1.34_{\pm 0.02}$ \\
BALF-F-0.8 & $-19.22_{\pm 0.00}$ & $-19.79_{\pm 0.00}$ & $-0.19_{\pm 0.01}$ \\
BALF-P-0.8 & $-19.36_{\pm 0.00}$ & $-19.94_{\pm 0.00}$ & $-0.17_{\pm 0.01}$ \\
\bottomrule
\end{NiceTabular*}
\end{subtable}
\hfill
\begin{subtable}[t]{0.49\textwidth}
\caption{MobileNetV2}
\centering
\begin{NiceTabular*}{\linewidth}{@{\extracolsep{\fill}}p{2.1cm}ccc@{}}
\CodeBefore
  \rectanglecolor{LightRoyalBlue}{7-1}{11-4}
\Body
\toprule
Method & $\Delta$FLOPs\% $\downarrow$ & $\Delta$Params\% $\downarrow$ & $\Delta$Top-1 pp $\uparrow$ \\
\midrule
SVD-NAS     & $-15.09$ & $-9.00$ & $-12.54$ \\
ALDS        & $-2.62$ & $-37.61$ & $-16.95$ \\
LR-S2       & $-3.81$ & $-6.24$ & $-17.46$ \\
L2+REST     & $-$ & $-5.00$  & $-4.42$  \\
L2+REST     & $-$ & $-10.00$ & $-18.47$ \\
BALF-P-0.7  & $-2.07_{\pm 0.01}$ & $-29.72_{\pm 0.02}$ & $-28.08_{\pm 0.50}$ \\
\mbox{BALF-P-0.75} & $-1.13_{\pm 0.01}$ & $-24.78_{\pm 0.02}$ & $-6.17_{\pm 0.05}$ \\
BALF-P-0.8  & $-0.23_{\pm 0.00}$ & $-19.80_{\pm 0.00}$ & $-0.61_{\pm 0.03}$ \\
BALF-P-0.9  & $-0.12_{\pm 0.00}$ & $-10.69_{\pm 0.05}$ & $-0.23_{\pm 0.01}$ \\
\mbox{BALF-F-0.975} & $-2.51_{\pm 0.00}$ & $-1.55_{\pm 0.00}$ & $-4.89_{\pm 0.02}$ \\
\bottomrule
\end{NiceTabular*}
\end{subtable}
\end{table*}
}

\begin{figure*}[h]
\centering

\begin{minipage}[t]{0.52\textwidth}
\centering
\vspace{0pt}
\small
\renewcommand{\arraystretch}{0.95}
\setlength{\tabcolsep}{3pt}
\begin{NiceTabular}{p{2.0cm}ccc}
\CodeBefore
  \rectanglecolor{LightRoyalBlue}{8-1}{14-4}
\Body
\toprule
Method & $\Delta$FLOPs\% $\downarrow$ & $\Delta$Params\% $\downarrow$ & $\Delta$Top-1 pp $\uparrow$ \\
\midrule
ALDS    & $-$ & $-23.71$ & $-1.50$  \\
ALDS    & $-$ & $-30.05$ & $-3.37$  \\
ALDS    & $-$ & $-50.00$ & $-25.22$ \\

FVRCP   & $-19.84$ & $-$ & $-2.50$   \\
FVRCP   & $-29.51$ & $-$ & $-5.83$  \\
FVRCP   & $-49.35$ & $-$ & $-24.86$ \\

BALF-F-0.5 & $-50.02_{\pm 0.00}$ & $-49.53_{\pm 0.08}$ & $-24.51_{\pm 0.23}$ \\
BALF-P-0.5 & $-24.75_{\pm 0.00}$ & $-49.74_{\pm 0.00}$ & $-12.94_{\pm 0.26}$ \\
BALF-F-0.7 & $-30.47_{\pm 0.19}$ & $-29.06_{\pm 0.03}$ & $-5.91_{\pm 0.24}$ \\
BALF-P-0.7 & $-12.78_{\pm 0.02}$ & $-29.91_{\pm 0.05}$ & $-2.52_{\pm 0.04}$ \\
BALF-F-0.8 & $-20.03_{\pm 0.00}$ & $-17.90_{\pm 0.00}$ & $-2.25_{\pm 0.05}$ \\
BALF-P-0.8 & $-9.11_{\pm 0.05}$ & $-21.31_{\pm 0.11}$ & $-1.24_{\pm 0.05}$ \\

\bottomrule
\end{NiceTabular}
\end{minipage}
\hfill
\begin{minipage}[t]{0.45\textwidth}
\centering
\vspace{0pt}
\includegraphics[width=\linewidth]{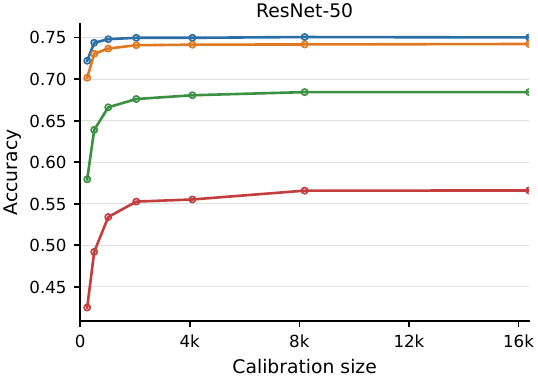}

\includegraphics{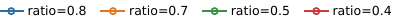}
\end{minipage}

\vspace{0.4em}

\begin{minipage}[t]{0.52\textwidth}
\captionof{table}{Comparison with compression methods on ResNet-20 (CIFAR-10). Notation follows \cref{tab:results_comparison}.}
\label{tab:resnet20_results_table}
\end{minipage}
\hfill
\begin{minipage}[t]{0.46\textwidth}
\captionof{figure}{Effect of calibration dataset size on ResNet-50 compression.}
\label{fig:calib_size_sweep_resnet50_main}
\end{minipage}

\end{figure*}

\paragraph{Results against factorization baselines.} \Cref{fig:posttraining-results} summarizes the complexity--accuracy tradeoffs across the different settings. Generally, activation-aware SVD yields consistent gains over plain SVD, with the exact gap varying by configuration. Moreover, our rank allocator yields further accuracy gains in most cases (but not every case) with respect to the optimized metric (FLOPs or parameter counts). It is important to note, however, that optimizing for one complexity metric can degrade the other. For example, in some cases, setting a parameter count objective yields substantial gains in parameter count but leads to worse FLOP reduction. For ViTs, the FLOP and parameter count curves are nearly identical. This is because, ignoring operations outside the main linear layers (like attention), compressing a linear layer reduces FLOPs by the same proportion as it reduces the number of parameters. The weakest results are achieved on MobileNetV2, which is a notably challenging case for factorization: activation-aware SVD yields relatively substantial improvements in the parameter count setting, with the allocator providing a slight additional edge, but FLOP reduction remains difficult.

\paragraph{Comparison with other fine-tuning-free compression approaches. } Additionally, we compare BALF with other structured compression approaches in the fine-tuning-free regime, with the baselines for each model determined by the availability of reported results. These include the pruning methods DFPC \citep{narshana2023dfpc}, IterTVSPrune (ITVSP) \citep{murti2023tvsprune}, IFM \citep{chen2024going}, FVRCP \citep{FVRCP}, DC-ViT \citep{zhang2024dense}, and PRACTISE \citep{wang2023practicalnetworkaccelerationtiny}; the factorization methods ALDS \citep{liebenwein2021compressing}, SVD-NAS \citep{yu2022svdnascouplinglowrankapproximation} and its LR-S2 baseline, and the methods studied in \citet{kalle2025distributionaware}; the token-merging method GTP \citep{xu2024gtpvitefficientvisiontransformers}; and the pruning-restoration method L2+REST \citep{lee2025trainingfreerestorationprunedneural}. For \citet{kalle2025distributionaware}, T-ALS and CP-ALS denote the plain alternating least-squares variants, while the -S variants denote their data-aware counterparts. We take results from their respective papers with the following exceptions. DeiT-B/16 results for PRACTISE are those reported by \citet{zhang2024dense}, and the no-fine-tuning MobileNetV2 results for ALDS are those reported by \citet{yu2022svdnascouplinglowrankapproximation}. For L2+REST, which reports the fraction of removed filters instead of parameters, we use that fraction as an approximation of the parameter count reduction.

\paragraph{Results against other compression approaches.} \Cref{tab:results_comparison} summarizes the results for ImageNet-1K models, and \cref{tab:resnet20_results_table} includes the results for ResNet-20 on CIFAR-10. Overall, BALF achieves performance comparable to most of the methods considered. It outperforms existing methods in some situations, although it is outperformed in others. Note, however, that some of these methods are substantially more expensive (e.g., SVD-NAS requires a search procedure on the order of hours, whereas BALF runs in minutes).

\paragraph{Compression runtime.} On an RTX 2080 Ti GPU, compressing each ImageNet-1K model we tested takes under four minutes. The most expensive steps can be cached, reducing subsequent compression runs to seconds. We provide a benchmark and implementation details in \cref{sec:runtime_impl}.

\paragraph{Robustness under calibration--evaluation mismatch.}
We evaluate the robustness of BALF under distribution shift in the calibration data. We do so by sampling calibration data from shifted datasets: CIFAR-10-C \citep{hendrycks2018benchmarking} for ResNet-20, and ImageNet-C \citep{hendrycks2018benchmarking} and ImageNet-V2 \citep{pmlr-v97-recht19a} for ResNet-18. We then evaluate the compressed models as usual on the original evaluation datasets. We find that results vary depending on the setting. On ImageNet-C and CIFAR-10-C, the effect generally depends on the corruption type and intensity: some corruptions produce only marginal decreases in performance, while more extreme corruptions can substantially affect BALF. In contrast, ImageNet-V2 calibration data yields no degradation, suggesting that BALF is robust to natural calibration shifts. Full results are presented in \cref{sec:additional_experiments}.

\paragraph{Robustness to calibration dataset size.}
We evaluate BALF using different calibration dataset sizes. We find that fewer calibration samples often yield only marginally worse results than those reported in the main experiments (about 256 samples often suffice for CIFAR-10, while about 1024--2048 samples often suffice for ImageNet-1K), depending on the setting. For instance, \cref{fig:calib_size_sweep_resnet50_main} shows a sweep on ResNet-50. Other results, following similar trends, are presented in \cref{sec:additional_experiments}.

\section{Discussion and Conclusion}

\paragraph{Limitations and future work.}
BALF inherits some limitations from low-rank factorization. For example, depthwise convolutional layers cannot be compressed within our framework because the only available rank is $P=1$. More generally, the $(\ExprOut,\ExprInp,\ExprParam)$ framework may not accommodate every layer type. In addition, although BALF reduces theoretical FLOPs and parameter counts, realizing end-to-end speedups can require efficient low-rank operators; for some configurations, taking advantage of the resulting low-rank structure remains challenging in practice (see \cref{subsec:practical_speedup}).

Our rank allocator is also approximate. In particular, it can undershoot the target budget in specific cases (see \cref{subsec:lagrangian}). Although we did not observe this behavior in our experiments, it can occur theoretically for networks with very sharp spectra, for example as a result of low-rank regularization \citep{ghosh2025qr}. Extending the allocator to support additional constraint types and concurrent constraints is also an interesting direction for future work. Another is to combine activation-aware projections with more expensive search methods, such as SVD-NAS, which could improve accuracy at the expense of compression time.

Finally, while BALF performs well on the vision models considered in this work, preliminary experiments on LLMs showed degradation relative to reported results from state-of-the-art factorization methods such as SVD-LLM. Since SVD-LLM-like projections are nearly equivalent to ours in practice, this suggests that our rank allocator may not be suitable in all settings.

\paragraph{Conclusion.}
We introduced BALF, a fine-tuning-free compression pipeline that generalizes activation-aware low-rank factorization beyond linear layers through our $(\ExprOut,\ExprInp,\ExprParam)$-expressible layer framework. Building on this framework, we provided a more general whitening-based formulation that covers a broader class of whitening matrices and yields an optimal activation-aware projection even under rank-deficient activations. We additionally proposed a rank allocator that selects per-layer ranks under global FLOP or parameter count budgets. Empirically, BALF generally improves the complexity--accuracy tradeoff over SVD-based baselines across CNNs and vision transformers and achieves competitive performance against other fine-tuning-free compression methods.

\acknowledgements

\subsubsection*{AI Use Statement}
We used AI to help identify related work, including potential comparison baselines, in addition to manual search. All cited works and reported comparisons were read and selected by us, and AI only assisted with search. The main ideas and their development are our own, but we also used AI to discuss research ideas (including identifying connections with other works) and assist with technical claims. More concretely, the most impactful suggestion from AI was that knapsack problems can be solved approximately. AI also helped catch and fix small errors in preliminary versions of our proposed whitening-based formulation and the accompanying claims. AI was also used to support code development (mostly limited to boilerplate experimental code and preliminary code versions for quick exploration) and to proofread and polish the writing.

\bibliography{main}
\bibliographystyle{tmlr}

\newpage
\appendix
\raggedbottom
\onecolumn

\crefalias{section}{appendix}
\crefalias{subsection}{appendix}
\crefalias{subsubsection}{appendix}

\section*{Appendix Overview}
This appendix is organized as follows:
\begin{itemize}
  \item \Cref{sec:thorough_decomposition} expands on the factorization scheme and provides parameter and FLOP counts for layers before and after low-rank factorization. It includes results for fully connected, ungrouped convolutional, and grouped convolutional layers, and motivates the use of a shared per-group rank for the latter.
  \item \Cref{sec:appendix_uwm} details the construction of uncentered whitening matrices via eigendecomposition, discusses numerical choices and fallbacks, and describes the incremental estimation of second moments and other details.
  \item \Cref{sec:rank_allocator_appendix} discusses the specifics of our rank allocator: it specifies the cost model, and presents the previously mentioned Lagrangian relaxation, together with complexity analysis and experimental results on its behavior.
  \item \Cref{sec:nn_distortion_bounds} states and proves the formal counterpart of \cref{prop:nn_bounds_prop_simple} (discussed in the main text), including simplified forms for special cases.
  \item \Cref{sec:experimental_details} documents the experimental details of our main experiments, including datasets, models, preprocessing, compression protocols, and other important details.
  \item \Cref{sec:additional_experiments} includes additional experiments that answer secondary questions, including reports on calibration dataset size ablations, robustness under distribution shift, and practical speedups.
  \item \Cref{sec:runtime_impl} describes the end-to-end practical implementation of BALF at a high level and provides timing and peak memory measurements of its different parts.
  \item \Cref{sec:post_hoc_info_geometry} provides a brief analysis of BALF through the lens of \citet{shumaylov2025informationgeometryiterativeoptimization}.
  \item \Cref{sec:appendix_proofs} collects complete proofs for results deferred from the main text, namely \cref{prop:general_activation_aware_distortion,prop:optimality} and auxiliary lemmas.
\end{itemize}

\section{Decomposing the Layers and Complexity After Factorization}
\label{sec:thorough_decomposition}

In this section, we provide a more thorough discussion of how layers are decomposed. We also provide the parameter counts and FLOP counts before and after decomposition (these are used as cost functions in our rank allocator). Given a layer and a rank $P$, our scheme has the same complexity as basic SVD truncation to rank $P$; the difference lies in the (expected) quality of the outputs of the factorized layer.  We omit biases for simplicity, as these remain constant in the original and factorized layers. $B$ denotes the batch size. Strictly speaking, the FLOP counts reported below should all be doubled (as it stands, we count each multiply-accumulate as one FLOP), but we omit the doubling factor for readability.

\subsection{Fully Connected Layers}
Let the weight be $\mW \in \mathbb{R}^{I \times O}$. For $\mX \in \Rspace{B, I}$, the original layer computes
\[
f(\mX; \mW) = \mX \mW.
\]

It has a total of $I O$ parameters, and takes $B I O$ FLOPs.

\paragraph{After decomposition.} Given a rank $P$, we factor the weights into $\mW_{0} \in \Rspace{I, P}$ and $\mW_{1} \in \Rspace{P, O}$. The decomposed layer becomes
\[
f(\mX; \mW_{0}, \mW_{1}) = (\mX \mW_{0}) \mW_{1}.
\]

The total storage needed is the sum of the storage for each matrix, that is, $P (I + O)$, while the number of FLOPs is $B P (I + O)$ (the computation order is algebraically irrelevant but matters for efficiency).

\subsection{2D Convolutional Layers}

Let the kernel be $\tW \in \Rspace{C_o, C_i, H_k, W_k}$ and the (per-item) output spatial size be $H_o \!\times\! W_o$. For an input
$\tX \in \Rspace{B, C_i, H_i, W_i}$, the original layer computes
\[
f(\tX; \tW) = \tX * \tW,
\]
where $*$ denotes 2D convolution. The total number of parameters is $C_o C_i H_k W_k$, and the number of FLOPs is $B H_o W_o C_o C_i H_k W_k$.

\paragraph{After decomposition.}
Choose a rank $P$. We factor the kernel into
\[
\tW_{0} \in \Rspace{P, C_i, H_k, W_k}
\quad\text{and}\quad
\tW_{1} \in \Rspace{C_o, P, 1, 1},
\]
and the layer computes
\[
f(\tX; \tW_{0}, \tW_{1}) = \big(\tX * \tW_{0}\big) * \tW_{1}.
\]
The total number of parameters is $P (C_i H_k W_k + C_o)$, and the FLOPs are
$B H_o W_o P (C_i H_k W_k + C_o)$. Note that $H_o$ and $W_o$ subsume the additional convolution hyperparameters (padding, strides, dilation, etc.).

``Regular'' convolutions are a special case of grouped convolutions, but we include a separate discussion for the convenience of the reader. We will now discuss the general case.

\subsection{Grouped 2D Convolutional Layers}\label{subsec:grouped_conv_factors}

Let $G$ be the number of groups (assume $C_i$ and $C_o$ are divisible by $G$, as is generally required in modern frameworks). The grouped kernel has shape
\[
\tW \in \Rspace{C_o, C_i / G, H_k, W_k}.
\]
The original grouped convolution computes
\[
f(\tX; \tW) = \tX \;\mathbin{*_{G}}\; \tW,
\]
with a total of $C_o \frac{C_i}{G} H_k W_k$  parameters, and a total of $B H_o W_o C_o \frac{C_i}{G} H_k W_k$ FLOPs.

\paragraph{After decomposition. } Given the per-group rank $P$, we decompose the weights into
\[
\tW_{0} \in \Rspace{P G, \, C_i / G, \, H_k, \, W_k}
\quad\text{and}\quad
\tW_{1} \in \Rspace{C_o, \, P, \, 1, 1},
\]
and the layer computes two grouped convolutions in sequence:
\[
f(\tX; \tW_{0}, \tW_{1}) = \big(\tX \;\mathbin{*_{G}}\; \tW_{0}\big) \;\mathbin{*_{G}}\; \tW_{1}.
\]
The total parameter storage across all groups is $P (C_i H_k W_k + C_o)$, and the FLOPs are $B H_o W_o P ( C_i H_k W_k + C_o)$. The intermediate feature map has $PG$ channels in total---$P$ per group---but the parameter and FLOP counts simplify to the expressions above.

\paragraph{Why share group ranks?} As noted above, the decomposition with the same rank per group results in two sequential (grouped) convolutions. Having different ranks per group would mean that, implementation-wise, we would need $G$ separate convolutions of different channel sizes. Performing all $G$ convolutions in parallel would be suboptimal, as the largest convolution would make the rest wait. This, among other reasons, would make it difficult to take advantage of having different ranks per group.

\section{Uncentered Whitening Matrices}\label{sec:appendix_uwm}

\subsection{Formal Definition and Properties}\label{sec:app_formal_uwm}

We first give the exact definition used throughout the text.

\begin{definition}\label{def:uwm}
    We define an uncentered whitening matrix (UWM) $\mM \in \Rspace{I, I}$ for the data $\mX \in \Rspace{N, I}$ as a matrix such that $\mM^\top \mX^\top \mX \mM = N \overline{\mI}_R$, where $\overline{\mI}_{R} \in \Rspace{I, I}$ is the identity on the first $R$ coordinates and zero elsewhere, and $R$ is the rank of $\mX$. We further restrict the definition by requiring that $\operatorname{row}(\mX) = \operatorname{range}(\mM)$.
\end{definition}

The following two properties hold, with notation as in \cref{def:uwm}.

\begin{proposition}
    The last $I - R$ columns of a UWM $\mM$ are zero.
\end{proposition}
\begin{proof}
    By our definition, $\mM^\top \mX^\top \mX \mM = N \overline{\mI}_R$, which means that the last $I - R$ columns of $\mX \mM$ are zero, i.e., the last $I - R$ columns of $\mM$ lie in $\operatorname{ker}(\mX)$. These columns also lie in $\operatorname{row}(\mX)$. Since $\operatorname{ker}(\mX) \cap \operatorname{row}(\mX) = \{\mathbf{0}\}$, these columns must be zero.
\end{proof}

\begin{proposition}
    It holds that $\mX = \mX \mM \mM^+$.
\end{proposition}
\begin{proof}
    Note that $\mM\mM^+$ is the orthogonal projector onto $\operatorname{range}(\mM)$. By definition, $\operatorname{range}(\mM) = \operatorname{row}(\mX)$. Hence, for any $\vu \in \operatorname{row}(\mX)$, we have $\mM\mM^+\vu = \vu$. It follows that $\mM\mM^+\mX^\top = \mX^\top$. By transposing and using the fact that orthogonal projectors are symmetric, $\mX = (\mM\mM^+\mX^\top)^\top = \mX \mM \mM^+$.
\end{proof}
\subsection{Obtaining UWMs in Practice}
We now provide more details about how the uncentered whitening matrices are obtained. We begin with a discussion of past work.

The naive option to obtain an uncentered whitening matrix is to use the Cholesky decomposition of the second-moment matrix, as explained in \citet{wang2025svdllm}. An issue is that we cannot compute it in the case of singular positive-semidefinite matrices. In their implementation,\footnote{\url{https://github.com/AIoT-MLSys-Lab/SVD-LLM}} they address this issue by adding a diagonal shift. Later, in \citet{wang2025svdllmv2optimizingsingular}, the authors note this issue and propose the use of an SVD-based whitening matrix, which can be used to support data residing in a subspace, although their analysis does not seem to cover this case (at the time of writing, the SVD-LLM V2 code has not been released). Nevertheless, they report slight end-to-end accuracy gains coming from the SVD-based whitening.

As noted in \cref{subsec:whitening_matrices}, we use an eigendecomposition-based scheme, implemented in practice through the $\mathsf{eigh}$ routine in PyTorch (a routine that obtains the eigenvalues and eigenvectors of a Hermitian (symmetric in the case of $\mathbb{R}$) matrix, and is usually faster than calling the corresponding generic eigendecomposition routine). For notational convenience, we assume the eigenpairs are ordered so that the positive eigenvalues appear first. In implementation, this ordering is unnecessary because the construction and subsequent computations are invariant to a simultaneous permutation of the whitening coordinates.

Recall that, for a single group $g$, with $\mX = \ExprInp(\tX)_g \in \Rspace{N, I}$, we compute
\[
[\mV, \mLambda] = \mathsf{eigh}(\frac{1}{N} \mX^\top \mX),
\]
and set
\[
\mM = \mV \mLambda^{+1/2}, \quad \mM^{+} = \mLambda^{1/2} \mV^\top,
\]
where
\[
\mV \mLambda \mV^\top = \frac{1}{N} \mX^\top \mX.
\]

It is easy to show that it satisfies our definition. Indeed,
\[
\mLambda^ {+1/2} \mV^\top \mX^\top \mX \mV \mLambda^ {+1/2} = N \mLambda^ {+1/2} \mV^\top \mV \mLambda \mV^\top \mV \mLambda^ {+1/2} = N \overline{\mI}_R,
\]
$R$ being the rank of $\mX$. Moreover, it is easy to see that $\operatorname{row}(\mX) = \operatorname{range}(\mM)$, as required in our definition (see \cref{sec:app_formal_uwm}). Additionally, the pseudoinverse of $\mLambda$ (a diagonal matrix) can be obtained efficiently by inverting each nonzero element of the diagonal, while leaving zeros as is (in practice, we use a small numerical threshold). This avoids additional numerical approximation errors when forming $\mM$. We do not perform any additional steps on the whitening matrices.

\paragraph{Practical considerations.} We use the linear algebra functions (with CUDA tensors) provided in the PyTorch library. During early experimentation, we used the MAGMA solver \citep{10.1177/10943420241261960}, which appeared to be more stable, but it was significantly slower than the default (cuSOLVER \citep{nvidia2025cusolver}), which was ultimately used in our experiments. Sporadically, the $\mathsf{eigh}$ routine is unable to converge. For that reason, we add a fallback to the SVD-based approach described below. In none of our experiments did the fallback also fail.

\paragraph{The alternative.} An SVD-based UWM, as proposed in SVD-LLM V2 \citep{wang2025svdllmv2optimizingsingular}, can be obtained in a similar way with the SVD of $\mX^\top\mX$. Although it is algebraically equivalent, it tends to be slower than our approach.

\paragraph{Accumulating the uncentered moments incrementally. } The naive way of obtaining $\frac{1}{N} \mX^\top \mX$ in practice is to use a single batch, and then cache activations by passing the batch through the network. However, for some networks with high-dimensional features and/or large batch sizes, the memory requirements might make the process expensive or impossible. A nice property (also mentioned in \citet{wang2025svdllm}, but not discussed extensively) is that we can use data batches to accumulate the second moments. Suppose we have $M$ batches, then,
\[
\frac{1}{N} \mX^\top \mX 
= \frac{1}{N} 
\begin{bmatrix}
\mX_{1,:}^\top & \cdots & \mX_{M,:}^\top \\
\end{bmatrix}
\begin{bmatrix}
\mX_{1,:} \\
\vdots \\
\mX_{M,:}
\end{bmatrix}
= \frac{1}{N}
\left(
\mX_{1,:}^\top \mX_{1,:} + \cdots + \mX_{M,:}^\top \mX_{M,:}
\right),
\]
where $N$ is the total outer dimension after $\ExprInp$ reshapes the activations (e.g., for convolutions, $N = B H_o W_o$). Here, the indexing denotes the selection of a sub-batch. We use this approach in our implementation.

\section{Rank Allocation Details}\label{sec:rank_allocator_appendix}

This section provides details about the rank allocation strategy discussed in \cref{subsec:selecting_rank_per_layer}.

Our objective is to select a list $\{P_l\}_{l=1}^{L}$ denoting the retained rank in each layer, with the objective of maximizing the global retained energy under some cost constraint. We can model this as a binary linear program in the form of a multiple-choice knapsack problem (for a review, see \citet{math13071097}), with decision variables $\{x_{l,t}\}_{l,t=1}^{L, U_l}$, where $x_{l,t} \in \{0, 1\}$ denotes whether, at layer $l$, $t$ is selected as the maximum retained rank, and $U_l$ denotes an upper bound on the rank (in our case, the number of singular values). Furthermore, by establishing a function $C_l(P_l)$ that returns the computational cost (e.g., in FLOPs or number of parameters) of keeping up to rank $P_l$ at layer $l$, we can ensure that the complexity of the solution is constrained to some user-set quantity. The binary program can be expressed as:
\begin{equation*}
\begin{aligned}
\max_{\{x_{l,t}\}} \quad & 
  \sum_{l=1}^L \sum_{t=1}^{U_l} E_l(t)\, x_{l,t} \\[0.3em]
\text{s.t.}\quad 
& \sum_{t=1}^{U_l} x_{l,t} = 1, && l=1,\dots,L, \\[-0.2em]
& \sum_{l=1}^L \sum_{t=1}^{U_l} C_l(t)\, x_{l,t} \le C_{\max}, \\[-0.2em]
& x_{l,t} \in \{0,1\}, && \forall\, l,t.
\end{aligned}
\end{equation*}

\paragraph{Are there other options for the objective function?} Recall that we use the total retained energy, which is normalized, as the objective function. Following the insights of \cref{prop:nn_bounds_prop_simple} and \cref{prop:nn_bounds_prop_complete} (although they are not tight for deeper networks), one might speculate that, with appropriate accounting of each layer's importance, we could achieve better results. During our initial experiments, we also evaluated unnormalized distortion as a metric in place of energy and explored layer-wise weightings based on statistics such as output variance and entropy. None of these alternatives improved performance over the method presented here; instead, they increased complexity and posed additional computational overhead and implementation challenges.

\paragraph{Measuring the cost.} Our allocator makes use of a cost function. We refer the reader to \cref{sec:thorough_decomposition} for the discussion on the cost of regular and factorized layers, both in terms of FLOPs and in terms of the number of parameters. It is clear that not all $P_l$ selections result in fewer parameters or FLOPs. Given $P_l$ and a criterion, we decompose the layer only when we achieve gains in terms of that criterion (e.g., a reduction in FLOPs or parameters); see \cref{alg:balf_overview}. To reflect this in the cost function, we define it as a piecewise linear function, where the first region is monotonically increasing and represents the region where decomposing the layer results in compression, and the second remains constant, denoting that the action of ``keeping rank $P_l$'' will have the cost of the original layer. Summarizing, we define
\[
C_l(P_l) = \min \{C_l^{\mathrm{orig}}, C_l^{\mathrm{decomp}}(P_l) \}.
\]

We emphasize that this step is important so as to inform the solver that we will not factorize a layer if it is not beneficial.

\subsection{A Scalable Lagrangian Relaxation}
\label{subsec:lagrangian}

The aforementioned program can be solved with publicly available libraries (e.g., PuLP if using Python \citep{Dunning2011PuLP}). The issue is that solving it exactly is known to be NP-hard (see \citet{math13071097}). To overcome this, we employ a Lagrangian relaxation. Pseudocode defining our implementation is given in \cref{alg:lagrangian}. It accepts a hyperparameter $T$ denoting the number of iterations to use. For a more general discussion, see \citet[Chapter~11]{kellerer2004knapsack}. In particular, for a fixed multiplier $\lambda$, our per-layer update solves the relaxed subproblem in \citet[Equation~(11.14)]{kellerer2004knapsack}. For each layer, it selects the rank maximizing $E_l(t)-\lambda C_l(t)$. We then search over $\lambda$ by bisection, using the monotonicity of the induced total cost, and return the final allocation. As usual for such a relaxation-based heuristic, the returned primal allocation need not be globally optimal.

\begin{algorithm}[ht]
\caption{Lagrangian relaxation for our rank allocator}
\label{alg:lagrangian}
\begin{algorithmic}[1]
\Require
$\displaystyle L$ layers, each with $U_l$ choices;\\
  energy function $E_{l}(\cdot)$ and costs $C_{l}(\cdot)$; total budget $C_{\max}$;\\
  max iterations $T$ (e.g., \ $T\!=\!300$ in our experiments);\\
  Assumptions (in our problem instance): (1) the problem is feasible (at least one selection with total cost $\le C_{\max}$ exists); 
  (2) all costs and energies are nonnegative; 
  (3) within each layer $l$, costs and energies are nondecreasing.
\Ensure
  Selection $t_l$ for each layer $l$, one index $t_l\in\{1,\ldots,U_l\}$.
\State $\lambda_{\min}\gets 0,  \lambda_{\max}\gets 1$  \Comment{initialize dual bounds}

\For{each layer $l=1,\dots,L$}
  \State $t_l^{(0)} \gets \min \bigl(\arg\max_{t} E_l(t)\bigr)$ 
  \Comment{same as $\lambda=0$; break ties with smallest cost}
\EndFor
\State $C^{(0)}\gets\sum_l C_l(t_l^{(0)})$
\If{$C^{(0)} \le C_{\max}$}
  \State \Return $\{t_l^{(0)}\}$ 
  \Comment{the unconstrained max-energy selection is already feasible}
\EndIf
\Repeat \Comment{grow $\lambda_{\max}$ until feasibility at $\lambda_{\max}$}
  \For{each layer $l=1,\dots,L$}
    \State $t_l \gets \min \bigl(\arg\max_{t}\bigl(E_{l}(t) - \lambda_{\max}\,C_{l}(t)\bigr)\bigr)$  \Comment{break ties with smallest cost}
  \EndFor
  \State $C\gets\sum_{l} C_{l}(t_l)$
  \If{$C > C_{\max}$}
    \State $\lambda_{\max}\gets 2\lambda_{\max}$
  \EndIf
\Until{$C \le C_{\max}$} \Comment{now $\lambda_{\max}$ yields a feasible selection}

\For{$k=1$ \textbf{to} $T$} \Comment{bisection to shrink the feasible/infeasible bracket}
  \State $\lambda \gets (\lambda_{\min} + \lambda_{\max})/2$
  \For{each layer $l=1,\dots,L$}
    \State $t_l \gets \min \bigl(\arg\max_{t}\bigl(E_{l}(t) - \lambda\,C_{l}(t)\bigr)\bigr)$ \Comment{break ties with smallest cost}
  \EndFor
  \State $C\gets\sum_{l} C_{l}(t_l)$
  \If{$C > C_{\max}$}
    \State $\lambda_{\min} \gets \lambda$
  \Else
    \State $\lambda_{\max} \gets \lambda$ \Comment{keep smallest feasible $\lambda$}
  \EndIf
\EndFor  \Comment{bracket invariant: $C(\lambda_{\min})>C_{\max}$, $C(\lambda_{\max})\le C_{\max}$}

\For{each layer $l=1,\dots,L$}
  \State $t_l \gets \min \bigl(\arg\max_{t}\bigl(E_{l}(t) - \lambda_{\max}\,C_{l}(t)\bigr)\bigr)$
\EndFor
\State $C\gets\sum_{l} C_{l}(t_l)$
\State \Return $\{t_l\}$ \Comment{final feasible selection, good in practice}
\end{algorithmic}
\end{algorithm}

\paragraph{Complexity.} Each of the $T$ outer iterations scans all the candidates once, giving time $O(T\sum_l U_l)$. The initial doubling to find $\lambda_{\max}$ adds an $O(S\sum_l U_l)$ term, where $S$ is the number of doubling steps. Thus, the total time complexity is $O((T+ S + 1)\sum_l U_l)$ (in the case of a nontrivial solution).

\begin{figure}[t]
    \centering
    \begin{subfigure}{0.48\textwidth}
        \centering
        \includegraphics[width=\textwidth]{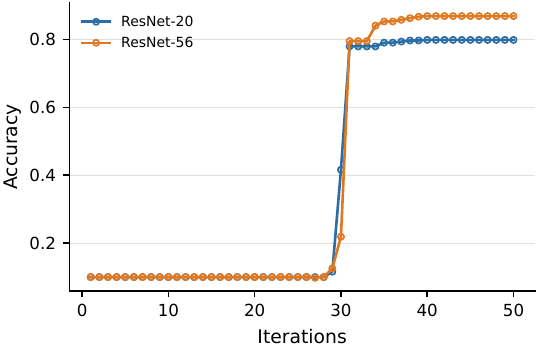}
    \end{subfigure}
    \hfill
    \begin{subfigure}{0.48\textwidth}
        \centering
        \includegraphics[width=\textwidth]{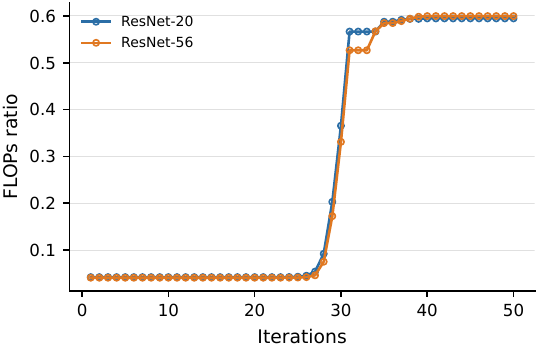}
    \end{subfigure}
    \caption{The effect of the number of iterations on CIFAR-10 models. We fix a compression target of 60\% of the original FLOPs. The horizontal axis denotes the number of iterations. The vertical axes denote the accuracy achieved (left) and the number of used FLOPs (right). The results stabilize with a relatively small number of iterations.
    }
    \label{fig:n_iters_results}
\end{figure}

\paragraph{Quality of the solutions.} The algorithm yields an approximate solution, but in every experiment we carried out, it produced good results and saturated (approximately) the target budget. It is, nevertheless, interesting to visualize how the algorithm behaves w.r.t.\ the number of iterations. \Cref{fig:n_iters_results} compares the number of iterations with two quantities of interest: model accuracy and the actual FLOPs used by the model (when using it with the objective of keeping 60\% of the FLOPs).

\paragraph{Undershooting.} We describe a specific situation in which our allocator can undershoot the budget. The allocator does not explicitly maximize budget utilization. Instead, it maximizes retained energy subject to an upper bound cost constraint. Consequently, the returned solution may use strictly less than the target budget. A simple case occurs when, for each layer, the retained energy curve saturates at some effective rank. Since the algorithm breaks ties in favor of the ``smallest-cost maximizer'', selection will not choose ranks larger than these minimal saturating ranks. If their total cost is already under the budget, the algorithm will not saturate the budget. Although we did not observe this behavior in our experiments, it can theoretically occur for networks with very sharp spectra, for example, as a result of low-rank regularization \citep{ghosh2025qr}.

\paragraph{Implementation details. } We implement the algorithm in Python, as even for larger models, it has a negligible runtime (less than 0.2 seconds in all our experiments, see \cref{sec:runtime_impl}). Moreover, although the energy and cost functions are represented in functional form in the algorithm description, in practice, they are precomputed as vectors for each layer.

\section{Bounding the Output Distortion of a Model}\label{sec:nn_distortion_bounds}

In this section, we provide a more thorough analysis of the claim in \cref{prop:nn_bounds_prop_simple}. We provide a more complete description and its proof. The result applies to sequential (possibly nonlinear) networks with $(\ExprOut, \ExprInp, \ExprParam)$-expressible layers (possibly with biases).

\begin{restatable}{proposition}{GeneralBoundsPropComplete}\label{prop:nn_bounds_prop_complete}
Define a sequential $L$-layer network
\[
\tX^{0} = \tX, \quad \tX^{l} = a_l\left(f_{l}(\tX^{l-1}; \tW^{l}) + \vb^l \right) = a_l \left(\ExprOut_{l}(\ExprInp_{l}(\tX^{l-1})\ExprParam_{l}(\tW^{l})) + \vb^l \right),
\]
where $a_l$ is an element-wise activation function with Lipschitz constant $A_l$. Moreover, let $B_l$ be the Lipschitz constant of $\ExprInp_{l}$ (note that $\ExprInp_{l}$ is linear and thus Lipschitz-continuous). We restrict our analysis to the case of ungrouped layers.
When approximating $\tW^{l}$ as a low-rank version, we denote the resulting parameter by $\widehat{\tW}{}^{l} = \AATrunc{P_l}{\tW^l}$, and the corresponding cumulatively distorted intermediate activations as $\widehat{\tX}{}^{l}$.

For each layer $l$, fix the retained rank and let
\[
\sqrt{\ell^{\mathrm{out}}_l}
=
\frac{1}{\sqrt{B}}\bigl\|\ f_{l}(\tX^{l-1}; \tW^{l}) - f_{l}(\tX^{l-1}; \widehat{\tW}{}^{l})\bigr\|_F,
\]
which reads as the (non-cumulative) error introduced by perturbing $\tW^l\to\widehat\tW{}^l$. Then, the total output distortion
\[
d^{(1:L)}
=\frac{1}{B}\bigl\|\tX^{L} - \widehat{\tX}{}^{L}\bigr\|^2_F
\]
satisfies the bound
\[
\sqrt{d^{(1:L)}}
\;\le\;
\sum_{l=1}^L
\Bigl[
A_l\sqrt{\ell^{\mathrm{out}}_l}\;\prod_{i=l+1}^L\|\ExprParam_i(\widehat{\tW}{}^{i})\|_2 A_iB_i
\Bigr].
\]

We follow the convention that the empty product is equal to 1.

\end{restatable}
\begin{proof}
For brevity, we set $\mW^l \equiv \ExprParam_l(\tW^l)$ (recall that we assume ungrouped layers).

Note that $\| \ExprOut_l(\cdot) \|_F = \|\cdot \|_F$, as $\ExprOut_l(\cdot)$ is a composition of permutations and reshapes. It is also true that $a_l \circ \ExprOut_l = \ExprOut_l \circ a_l$, as $a_l$ is element-wise.

Then:
\begin{align*}
\sqrt{B d^{(1:L)}}
&= \| \tX^{L} - \widehat{\tX}{}^{L} \|_F \\
&= \| a_L(f_{L}(\tX^{L-1}; \tW^{L}) + \vb^L) - a_L(f_{L}(\widehat{\tX}{}^{L-1}; \widehat{\tW}{}^{L}) + \vb^L) \|_F \\
&= \|  a_L(\ExprOut_L(\ExprInp_L(\tX^{L-1}) \mW^L) + \vb^L) 
     - a_L(\ExprOut_L(\ExprInp_L(\widehat{\tX}{}^{L-1}) \widehat{\mW}^L) + \vb^L) \|_F \\
&\leq A_L\|\ExprOut_L(\ExprInp_L(\tX^{L-1}) \mW^L) + \vb^L
     - \ExprOut_L(\ExprInp_L(\widehat{\tX}{}^{L-1}) \widehat{\mW}^L) - \vb^L \|_F \\
&= A_L\| \ExprInp_L(\tX^{L-1}) \mW^L
     - \ExprInp_L(\widehat{\tX}{}^{L-1}) \widehat{\mW}^L\|_F \\
&= A_L\| \ExprInp_L(\tX^{L-1})\mW^L  
     - \ExprInp_L(\tX^{L-1})\widehat{\mW}^L +  \ExprInp_L(\tX^{L-1}) \widehat{\mW}^L
     -  \ExprInp_L(\widehat{\tX}{}^{L-1}) \widehat{\mW}^L \|_F \\
&\leq A_L\| \ExprInp_L(\tX^{L-1})\mW^L  
     - \ExprInp_L(\tX^{L-1})\widehat{\mW}^L\|_F  +  A_L\|\ExprInp_L(\tX^{L-1}) \widehat{\mW}^L
     -  \ExprInp_L(\widehat{\tX}{}^{L-1}) \widehat{\mW}^L \|_F \\
&= A_L\sqrt{B\ell^{\mathrm{out}}_L} 
   + A_L\|\ExprInp_L(\tX^{L-1}) \widehat{\mW}^L
     -  \ExprInp_L(\widehat{\tX}{}^{L-1}) \widehat{\mW}^L \|_F \\
&\leq A_L\sqrt{B\ell^{\mathrm{out}}_L} 
   + \| \widehat{\mW}^L \|_2 A_L B_L \| \tX^{L-1} 
   - \widehat{\tX}{}^{L-1} \|_F \\
&= A_L \sqrt{B\ell^{\mathrm{out}}_L} 
   + A_L B_L\| \widehat{\mW}^L \|_2 \sqrt{B d^{(1:L - 1)}}.
\end{align*}

Unrolling the recursion and dividing by $\sqrt{B}$ completes the proof.

\end{proof}

\paragraph{Some special cases.} There are some interesting instances where the bound reduces to simpler expressions. For example, for fully connected layers, $B_l = 1$. Moreover, for 1-Lipschitz activation functions, such as ReLU \citep{10.5555/3104322.3104425}, $A_l = 1$. For instance, on MLPs with ReLU activations, the bound reduces to
\begin{align*}
    \sqrt{d^{(1:L)}} \leq \sum_{l = 1}^L \left[ \sqrt{\ell^{\mathrm{out}}_l} \prod_{i=l+1}^L \| \widehat{\mW}{}^i \|_2 \right].
\end{align*}

\paragraph{Looseness of the bound.} In deep networks, as noted in the main text, this bound is informative rather than practical, as it tends to become looser with depth; the main reason is that the bound involves products of spectral norms. There are some works, for instance \citet{pmlr-v70-cisse17a}, that aim to constrain the operator norms of the layers of neural networks. Such constraints would directly tighten our result, but they are not enforced in standard architectures.

\section{Experimental Details}\label{sec:experimental_details}

This section provides additional experimental details. All experiments were carried out using the PyTorch library.
We make our code public at \url{\githuburl}. To ensure reproducibility, we include (1) the library code that implements our methods, as well as the baselines, (2) the Python scripts used to carry out the experiments, (3) the Bash scripts used to call the Python scripts with the specific hyperparameters used, and (4) the dependencies and versions used.

\subsection{CIFAR-10 Experiments}

\paragraph{Models.} Since there are no widely adopted pre-trained CIFAR-10 ResNet model weights, we pre-train the ResNet-20 and ResNet-56 models ourselves, following standard practice. Our CIFAR-10 ResNet implementation is based on the \href{https://github.com/facebookarchive/fb.resnet.torch/blob/master/models/resnet.lua}{fb.resnet.torch} codebase, with the shortcut option set to B: the residual branch uses an identity mapping when the input and output shapes match, and otherwise applies a $1 \!\times\! 1$ convolution with the appropriate stride followed by batch normalization to match shapes. We use the default PyTorch parameter initialization for all layers.

We use SGD with momentum 0.9, weight decay $10^{-4}$, and a batch size of 128. The learning rate is initialized to 0.1 and reduced by a factor of 0.1 at epochs 100 and 150. We save the final model checkpoint at epoch 200. For data augmentation, we apply 4-pixel zero-padding followed by a random crop to $32 \!\times\! 32$ pixels, a random horizontal flip with a probability of 0.5, and per-channel normalization. Under these settings, the final test accuracies are 91.93\% for ResNet-20 and 93.18\% for ResNet-56.

\paragraph{Compression.} In order to obtain the $k$-calibration dataset, we pick $k$ random images uniformly from the training dataset. For all our main experiments, we use 1024 images.

All factorizable layers are considered for compression. For the CIFAR-10 ResNet models, this includes all the convolutional layers, including shortcut layers, and the final fully connected classifier.

\paragraph{Dataset processing.} We do not use preprocessing in our calibration dataset and test dataset, other than normalization, so as to mimic inference.

\subsection{ImageNet-1K Experiments}
\paragraph{Models. } For all the CNNs, we use the implementation provided by the $\mathsf{torchvision}$ library \citep{TorchVision}. We also use their $\mathsf{V1}$ checkpoints for all the models. For the experiments on vision transformers, we use the implementation and weights provided in the $\mathsf{timm}$ library \citep{Wightman_PyTorch_Image_Models}.

\paragraph{Compression.} In order to obtain the $k$-calibration dataset, we pick $k$ random images uniformly from the training dataset. For all our main experiments, we use 8192 images.

All factorizable layers are considered for compression. For the CNNs, this includes all their convolutional layers (including shortcuts) and the linear layers present in the classifier. For the vision transformers, this includes the initial convolutional layer, as well as the linear layers inside attention blocks and MLPs. For the attention blocks, we factorize the merged Q-K-V matrix (implemented in $\mathsf{timm}$ as a single linear layer).

\paragraph{Dataset processing.} Following standard inference practices, we first resize the shorter side of images to 248 for vision transformers (using bicubic interpolation) and to 256 (using bilinear interpolation) for the remaining models, and then center-crop them to $224\!\times\!224$. We then normalize them following the standard procedure. In particular, we use the default ImageNet-1K mean and standard deviations for all models except for the ViT-B/16, for which we follow the conventions of $\mathsf{timm}$ and use 0.5 as the mean and standard deviation for all channels. We do this for both the calibration dataset and the validation dataset.

\subsection{General Details}

For all our main experiments, we use 300 iterations for our rank allocator (see \cref{subsec:lagrangian}). For all the compression sweeps in our experiments section, we conduct a full sweep over the compression targets. For the experiments with our allocator and the Uniform baselines, we select compression ratios (w.r.t.\ parameter and FLOP counts) between 0.1 and 1.0 in steps of 0.05; additionally, we add 0.975. For the energy-based methods, we manually select an appropriate set of hyperparameters for each model so that they cover a suitable spectrum of compression ratios.

We follow standard practice when reporting compression results. For FLOP counts, we account for convolutional and fully connected layers, and for ViT models we also count attention FLOPs; we do not count secondary layers such as normalization layers. For FLOP targets, the allocator optimizes only over the FLOPs of layers eligible for factorization. Reported FLOP ratios are then computed after adding back the remaining counted but unoptimized operations. For parameter count results, we include all model parameters in the reported ratios, including parameters of layers that are not compressed. The allocator, however, only optimizes over parameters of layers eligible for factorization.

For some comparison results in the experimental section, we used WebPlotDigitizer\footnote{\url{https://automeris.io/wpd}} to extract numeric values from figures when the values were not reported numerically in tables. We took care to perform the extraction as accurately as possible.

\subsection{Computational Resources}
Earlier runs of the experiments in past manuscript versions were executed on a laptop with an RTX 2070 and a node with an RTX 4090. The experiments presented in the final version were each run on a single A100 40GB GPU unless noted otherwise. Preliminary experimentation that did not make it into the paper was conducted on a variety of hardware configurations, including an A100 40GB and other RTX GPUs.

\section{Additional Experimental Results}\label{sec:additional_experiments}

This section presents some additional experimental results that aim to answer secondary questions. 

\subsection{How Impactful is the Calibration Dataset Size?}

\begin{figure}[h]
  \centering

  \begin{minipage}{0.32\linewidth}
    \centering
    \includegraphics[width=\linewidth]{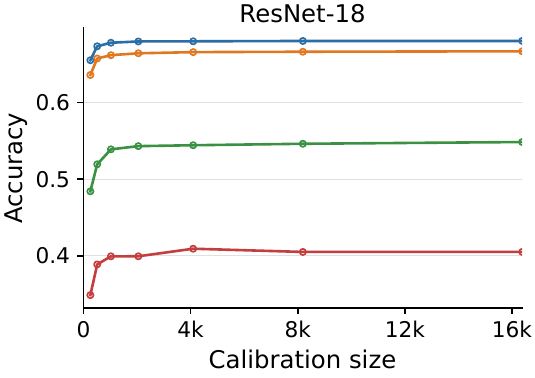}
  \end{minipage}\hfill
  \begin{minipage}{0.32\linewidth}
    \centering
    \includegraphics[width=\linewidth]{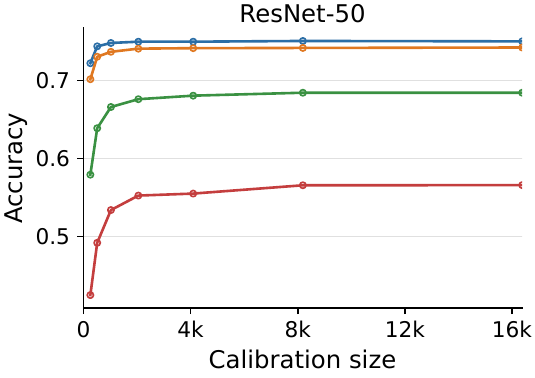}
  \end{minipage}\hfill
  \begin{minipage}{0.32\linewidth}
    \centering
    \includegraphics[width=\linewidth]{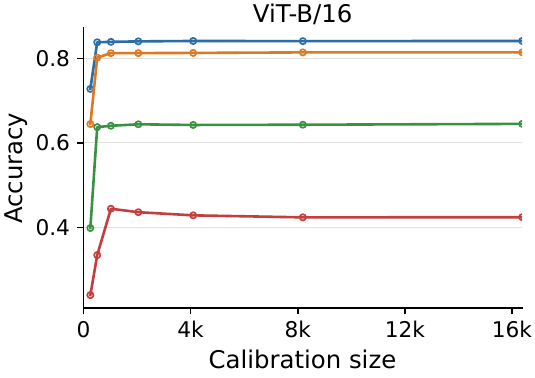}
  \end{minipage}\\[0.8em]

  \hfill
  \begin{minipage}{0.32\linewidth}
    \centering
    \includegraphics[width=\linewidth]{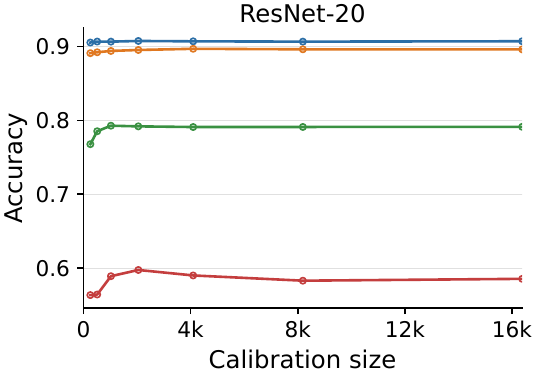}
  \end{minipage}\hfill
  \begin{minipage}{0.32\linewidth}
    \centering
    \includegraphics[width=\linewidth]{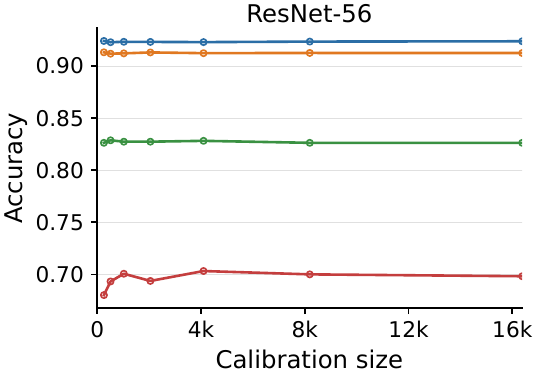}
  \end{minipage}
  \hfill\null

  \vspace{0.2em}
  \includegraphics{img/calib_size_sweep_legend.pdf}

  \caption{Calibration size vs. performance across models. Hyperparameters are fixed to those of the main experiments, with calibration sizes swept over $\{256,512,1024,2048,4096,8192,16384\}$. Ratios indicate the target parameter ratios passed to the allocator.}
  \label{fig:calib_size_sweep}
\end{figure}

Although we used 8192 calibration images for ImageNet-1K and 1024 for CIFAR-10, we noticed that moderate calibration dataset sizes worked well. Here, we carry out a sweep over different calibration dataset sizes and compression ratios with parameter compression targets. We use a ResNet-20 and ResNet-56 on CIFAR-10, and ResNet-18, ResNet-50 and ViT-B/16 on ImageNet-1K. \Cref{fig:calib_size_sweep} shows the different results. As shown, CIFAR-10 is not particularly sensitive to the size of the calibration dataset. On the other hand, the need for larger calibration sets is more apparent on ImageNet-1K. This need increases for the ResNet-50, and is exacerbated for higher compression ratios. However, beyond 1024--2048 images, more images yield only marginally better results.

\subsection{Robustness under Distribution Shift}

Our method relies on the computation of (uncentered) whitening matrices (over the training dataset in our main experiments) to estimate the optimal low-rank projections. We rely on the i.i.d.~assumption to expect transfer to unseen data. As shown over the training--test pairs on CIFAR-10 and ImageNet-1K, we obtain good results. In general, the i.i.d.~assumption is common (although seldom discussed) in methods that use calibration data.

Here, we evaluate BALF under calibration--evaluation mismatch. The base models are unchanged and are trained on the original training datasets. We then compute the calibration statistics using shifted data, but evaluate the resulting compressed models on the standard clean evaluation sets. The objective of these experiments is to test whether BALF remains effective when the calibration data used for compression differs from the data distribution used at evaluation time.

\begin{figure}[!p]
    \centering

    \begin{subfigure}[t]{0.32\linewidth}
        \centering
        \includegraphics[width=\linewidth]{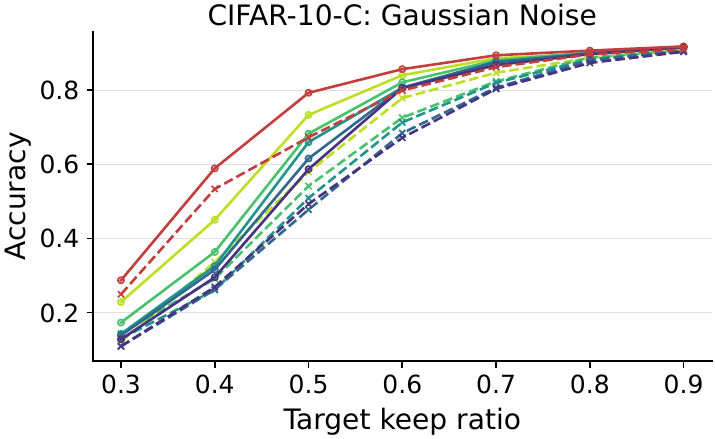}
        \label{fig:cifar10c_gaussian_noise}
    \end{subfigure}\hfill
    \begin{subfigure}[t]{0.32\linewidth}
        \centering
        \includegraphics[width=\linewidth]{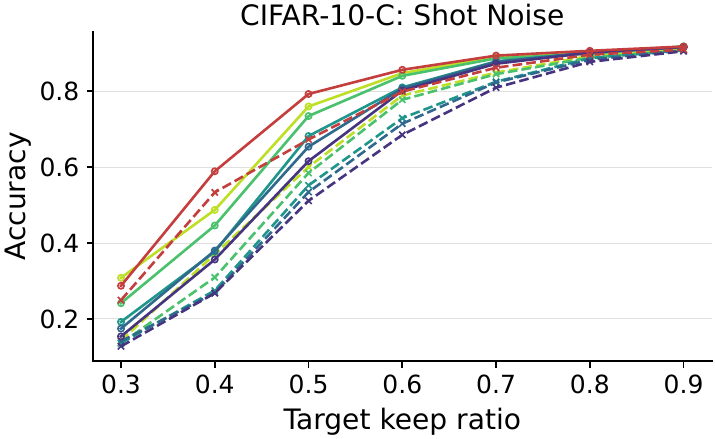}
        \label{fig:cifar10c_shot_noise}
    \end{subfigure}\hfill
    \begin{subfigure}[t]{0.32\linewidth}
        \centering
        \includegraphics[width=\linewidth]{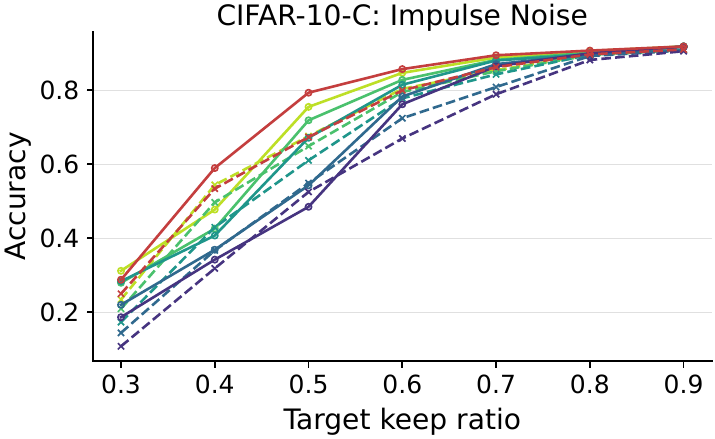}
        \label{fig:cifar10c_impulse_noise}
    \end{subfigure}

    \vspace{0.8em}

    \begin{subfigure}[t]{0.32\linewidth}
        \centering
        \includegraphics[width=\linewidth]{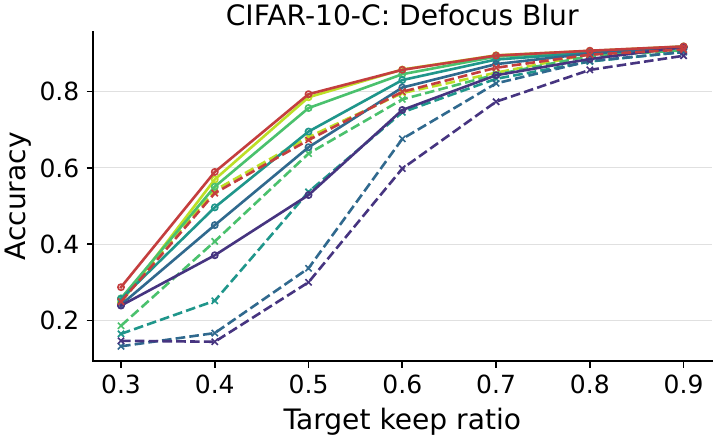}
        \label{fig:cifar10c_defocus_blur}
    \end{subfigure}\hfill
    \begin{subfigure}[t]{0.32\linewidth}
        \centering
        \includegraphics[width=\linewidth]{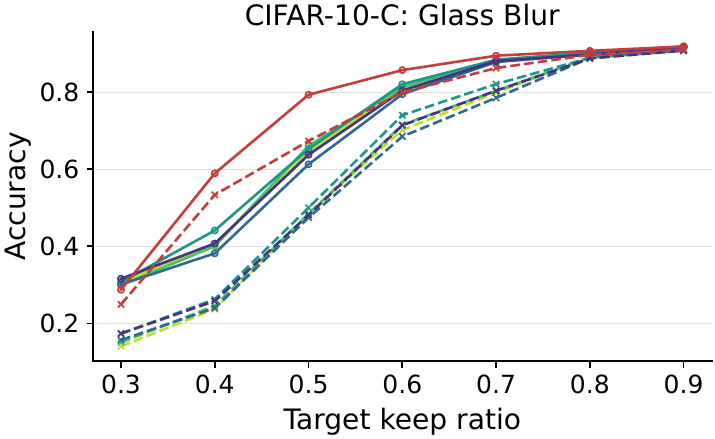}
        \label{fig:cifar10c_glass_blur}
    \end{subfigure}\hfill
    \begin{subfigure}[t]{0.32\linewidth}
        \centering
        \includegraphics[width=\linewidth]{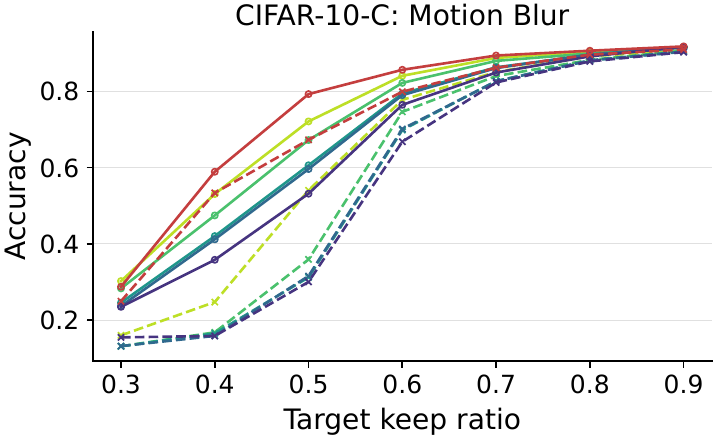}
        \label{fig:cifar10c_motion_blur}
    \end{subfigure}

    \vspace{0.8em}

    \begin{subfigure}[t]{0.32\linewidth}
        \centering
        \includegraphics[width=\linewidth]{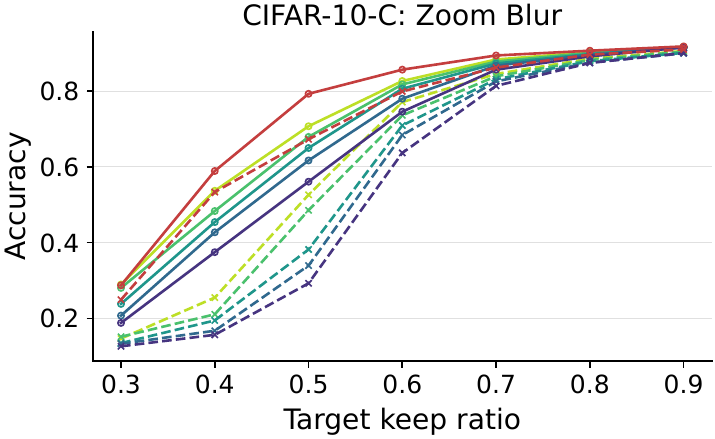}
        \label{fig:cifar10c_zoom_blur}
    \end{subfigure}\hfill
    \begin{subfigure}[t]{0.32\linewidth}
        \centering
        \includegraphics[width=\linewidth]{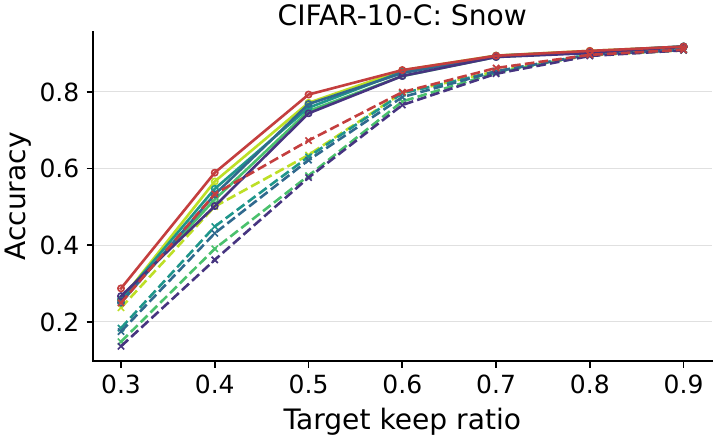}
        \label{fig:cifar10c_snow}
    \end{subfigure}\hfill
    \begin{subfigure}[t]{0.32\linewidth}
        \centering
        \includegraphics[width=\linewidth]{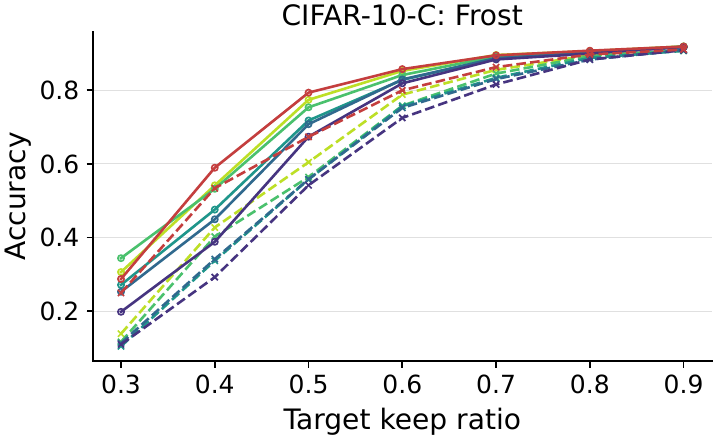}
        \label{fig:cifar10c_frost}
    \end{subfigure}

    \vspace{0.8em}

    \begin{subfigure}[t]{0.32\linewidth}
        \centering
        \includegraphics[width=\linewidth]{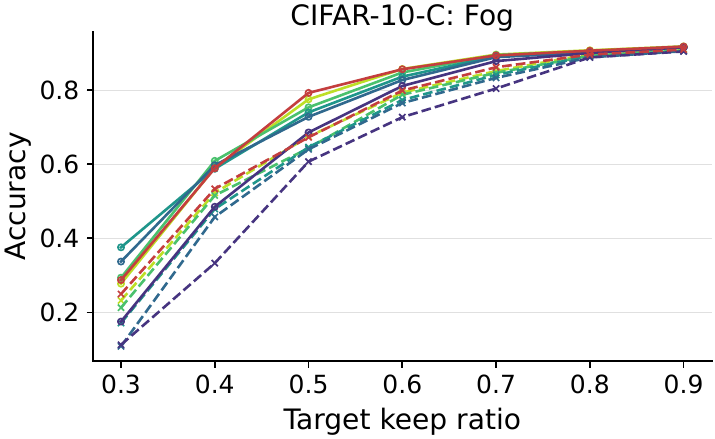}
        \label{fig:cifar10c_fog}
    \end{subfigure}\hfill
    \begin{subfigure}[t]{0.32\linewidth}
        \centering
        \includegraphics[width=\linewidth]{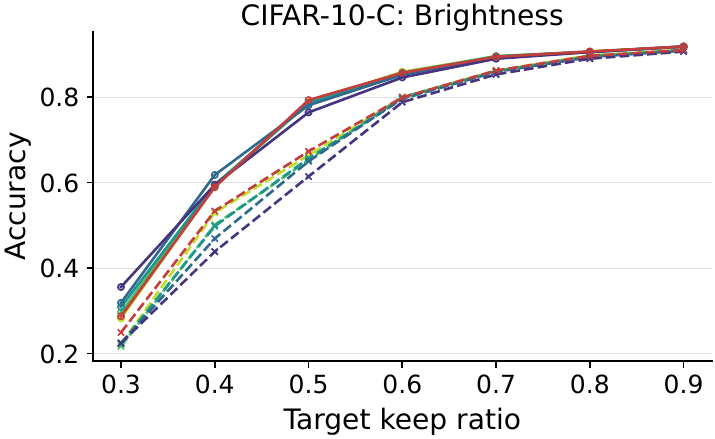}
        \label{fig:cifar10c_brightness}
    \end{subfigure}\hfill
    \begin{subfigure}[t]{0.32\linewidth}
        \centering
        \includegraphics[width=\linewidth]{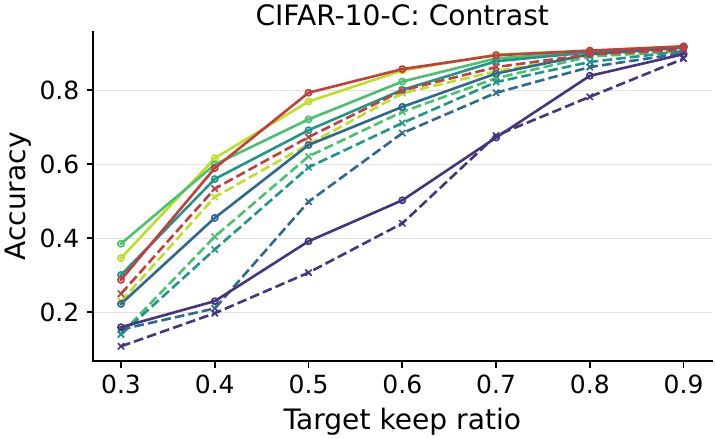}
        \label{fig:cifar10c_contrast}
    \end{subfigure}

    \vspace{0.8em}

    \begin{subfigure}[t]{0.32\linewidth}
        \centering
        \includegraphics[width=\linewidth]{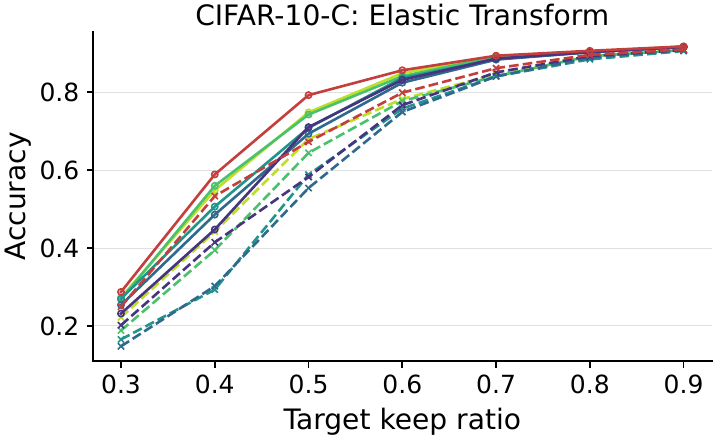}
        \label{fig:cifar10c_elastic_transform}
    \end{subfigure}\hfill
    \begin{subfigure}[t]{0.32\linewidth}
        \centering
        \includegraphics[width=\linewidth]{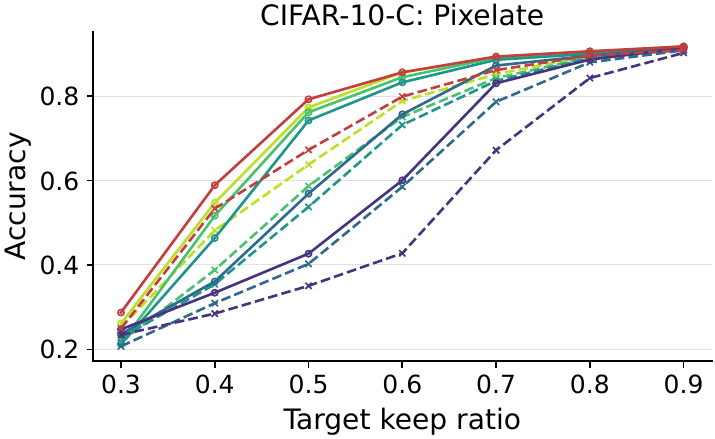}
        \label{fig:cifar10c_pixelate}
    \end{subfigure}\hfill
    \begin{subfigure}[t]{0.32\linewidth}
        \centering
        \includegraphics[width=\linewidth]{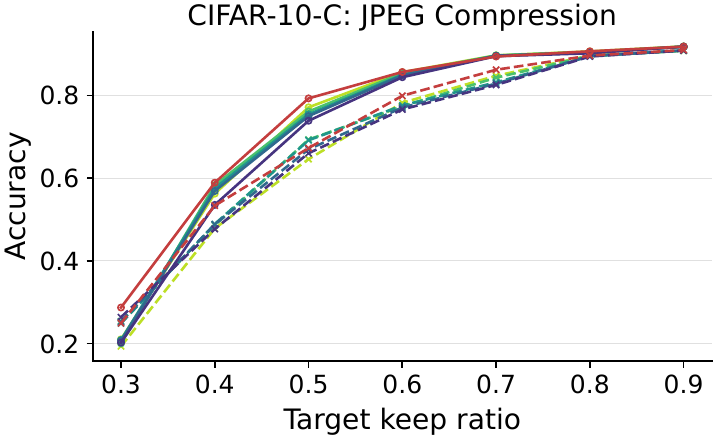}
        \label{fig:cifar10c_jpeg_compression}
    \end{subfigure}

    \vspace{0.2em}
    \includegraphics{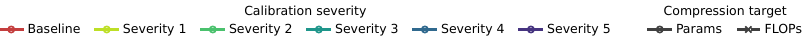}

   \caption{Compression--accuracy curves on ResNet-20 when calibration data are drawn from CIFAR-10-C and evaluation is performed on the clean CIFAR-10 test set. Each panel corresponds to a CIFAR-10-C corruption type, with curves shown across target keep ratios and calibration severities. The target keep ratio denotes the fraction of the original parameter count or FLOPs passed to the allocator as the budget. ``Baseline'' denotes calibration on the clean training set.}
    \label{fig:cifar10c_all_corruptions}
\end{figure}

\begin{figure}[ht]
    \centering

    \begin{subfigure}[t]{0.32\linewidth}
        \centering
        \includegraphics[width=\linewidth]{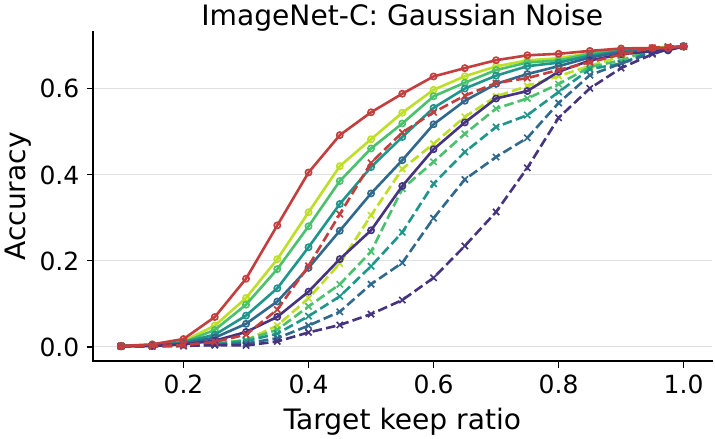}
        \label{fig:imagenetc_gaussian_noise}
    \end{subfigure}\hfill
    \begin{subfigure}[t]{0.32\linewidth}
        \centering
        \includegraphics[width=\linewidth]{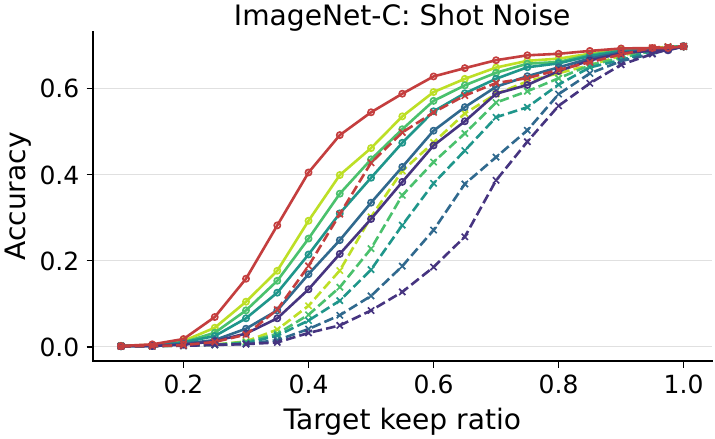}
        \label{fig:imagenetc_shot_noise}
    \end{subfigure}\hfill
    \begin{subfigure}[t]{0.32\linewidth}
        \centering
        \includegraphics[width=\linewidth]{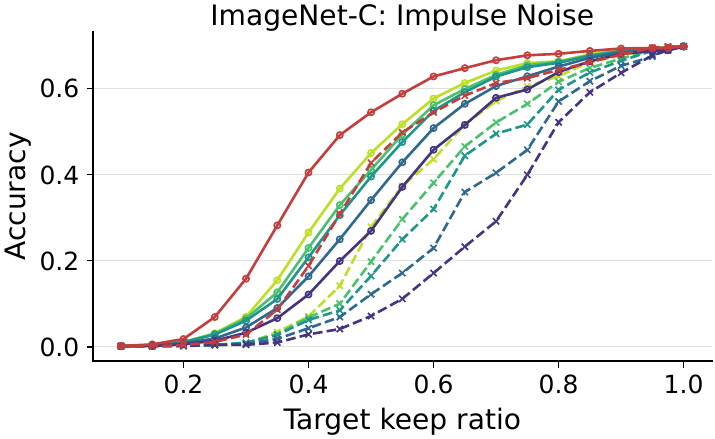}
        \label{fig:imagenetc_impulse_noise}
    \end{subfigure}

    \vspace{0.8em}

    \begin{subfigure}[t]{0.32\linewidth}
        \centering
        \includegraphics[width=\linewidth]{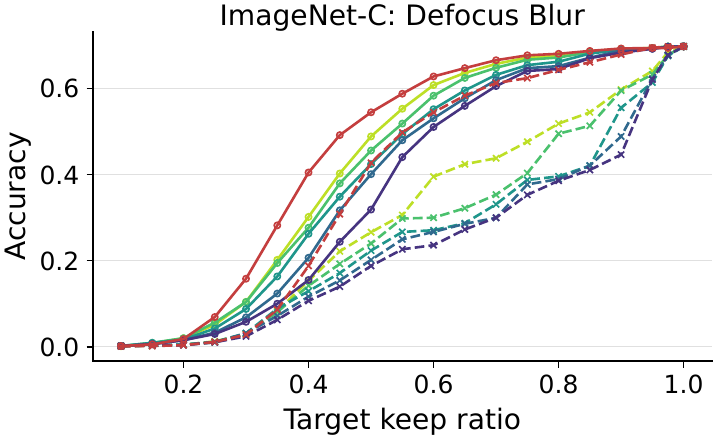}
        \label{fig:imagenetc_defocus_blur}
    \end{subfigure}\hfill
    \begin{subfigure}[t]{0.32\linewidth}
        \centering
        \includegraphics[width=\linewidth]{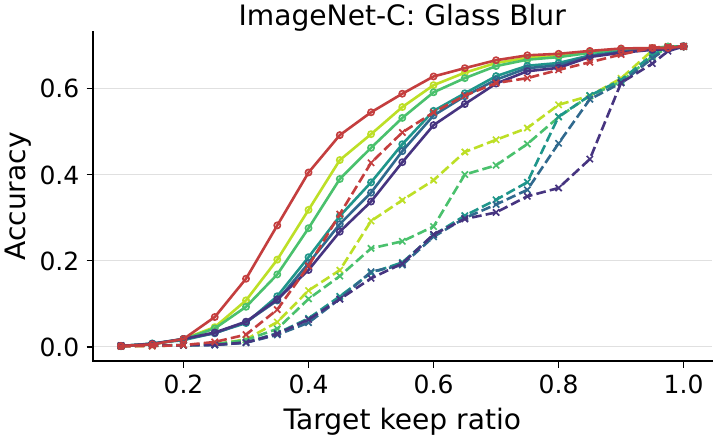}
        \label{fig:imagenetc_glass_blur}
    \end{subfigure}\hfill
    \begin{subfigure}[t]{0.32\linewidth}
        \centering
        \includegraphics[width=\linewidth]{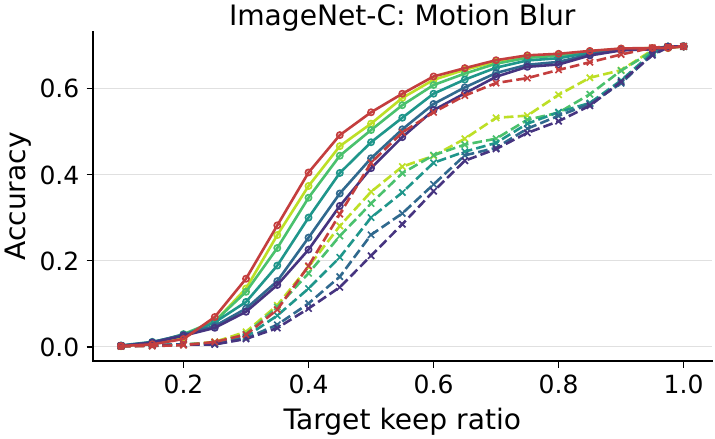}
        \label{fig:imagenetc_motion_blur}
    \end{subfigure}

    \vspace{0.8em}

    \begin{subfigure}[t]{0.32\linewidth}
        \centering
        \includegraphics[width=\linewidth]{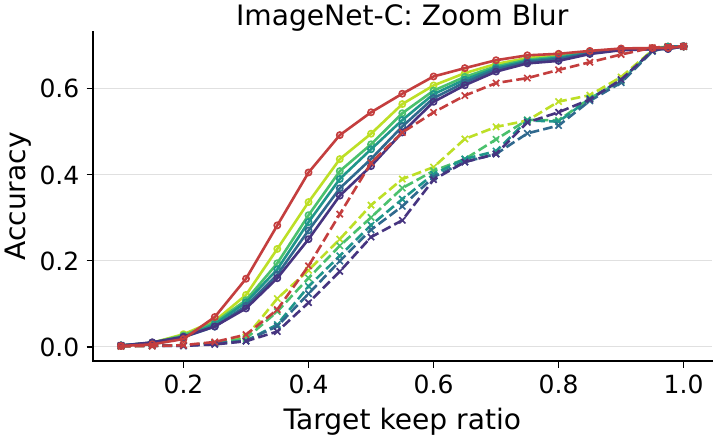}
        \label{fig:imagenetc_zoom_blur}
    \end{subfigure}\hfill
    \begin{subfigure}[t]{0.32\linewidth}
        \centering
        \includegraphics[width=\linewidth]{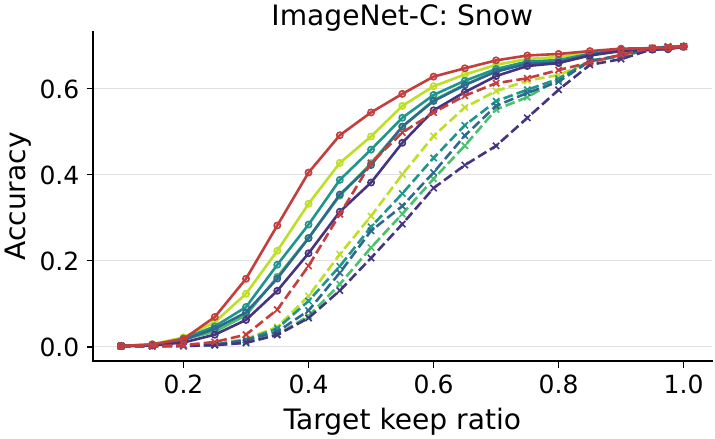}
        \label{fig:imagenetc_snow}
    \end{subfigure}\hfill
    \begin{subfigure}[t]{0.32\linewidth}
        \centering
        \includegraphics[width=\linewidth]{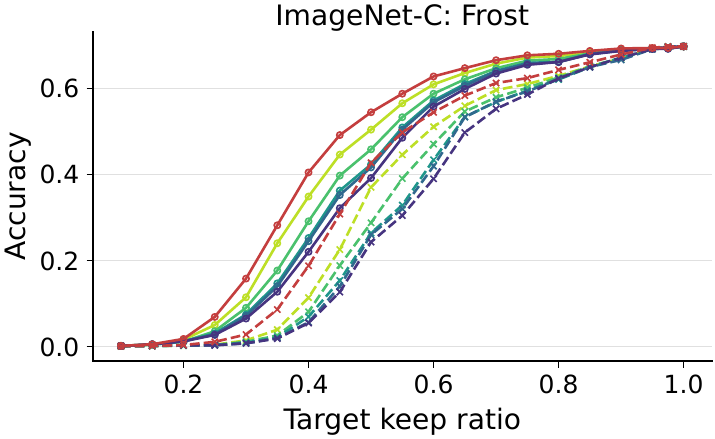}
        \label{fig:imagenetc_frost}
    \end{subfigure}

    \vspace{0.2em}
    \includegraphics{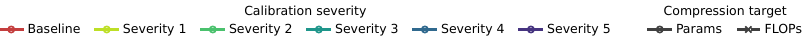}

    \caption{Compression--accuracy curves on ResNet-18 when calibration data are drawn from ImageNet-C and evaluation is performed on the standard ImageNet-1K validation set. Each panel corresponds to a ImageNet-C corruption type from the noise, blur, and selected weather categories. Notation as in \cref{fig:cifar10c_all_corruptions}.}
    \label{fig:imagenetc_all_corruptions_part1}
\end{figure}

\begin{figure}[ht]
    \centering

    \begin{subfigure}[t]{0.32\linewidth}
        \centering
        \includegraphics[width=\linewidth]{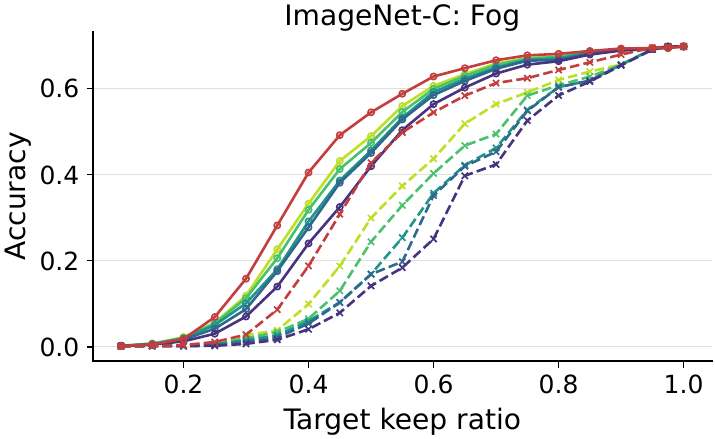}
        \label{fig:imagenetc_fog}
    \end{subfigure}\hfill
    \begin{subfigure}[t]{0.32\linewidth}
        \centering
        \includegraphics[width=\linewidth]{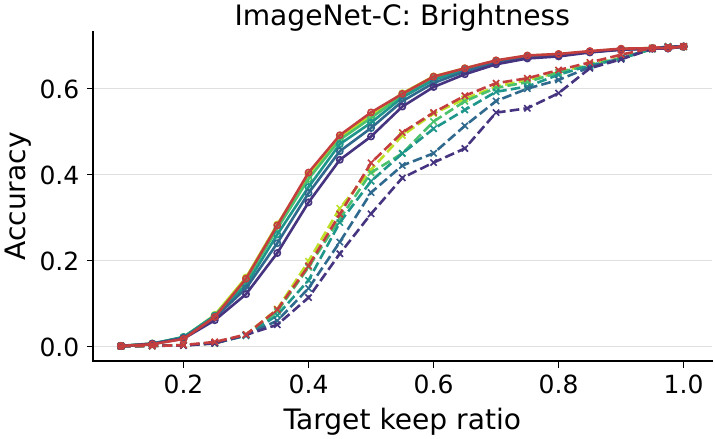}
        \label{fig:imagenetc_brightness}
    \end{subfigure}\hfill
    \begin{subfigure}[t]{0.32\linewidth}
        \centering
        \includegraphics[width=\linewidth]{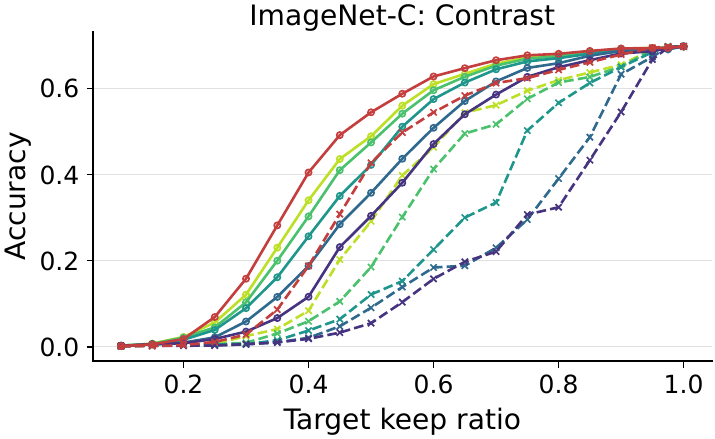}
        \label{fig:imagenetc_contrast}
    \end{subfigure}

    \vspace{0.8em}

    \begin{subfigure}[t]{0.32\linewidth}
        \centering
        \includegraphics[width=\linewidth]{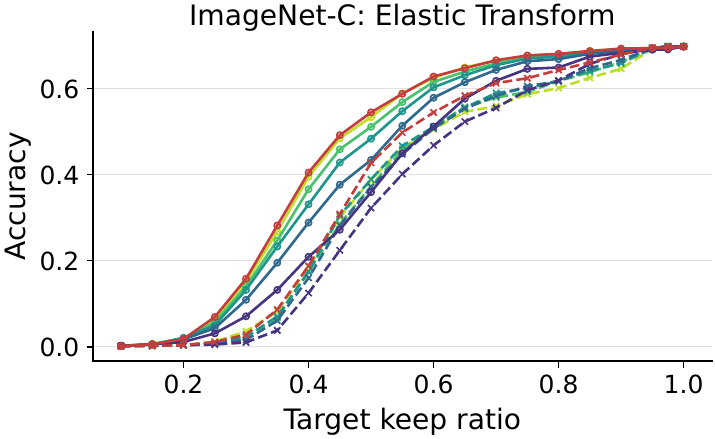}
        \label{fig:imagenetc_elastic_transform}
    \end{subfigure}\hfill
    \begin{subfigure}[t]{0.32\linewidth}
        \centering
        \includegraphics[width=\linewidth]{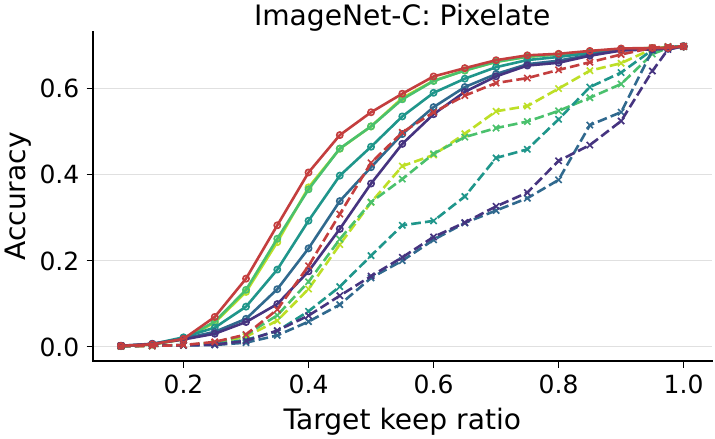}
        \label{fig:imagenetc_pixelate}
    \end{subfigure}\hfill
    \begin{subfigure}[t]{0.32\linewidth}
        \centering
        \includegraphics[width=\linewidth]{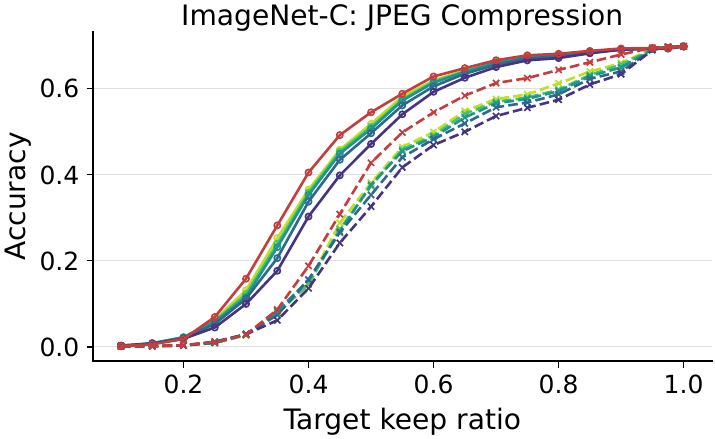}
        \label{fig:imagenetc_jpeg_compression}
    \end{subfigure}

    \vspace{0.8em}

    \begin{subfigure}[t]{0.32\linewidth}
        \centering
        \includegraphics[width=\linewidth]{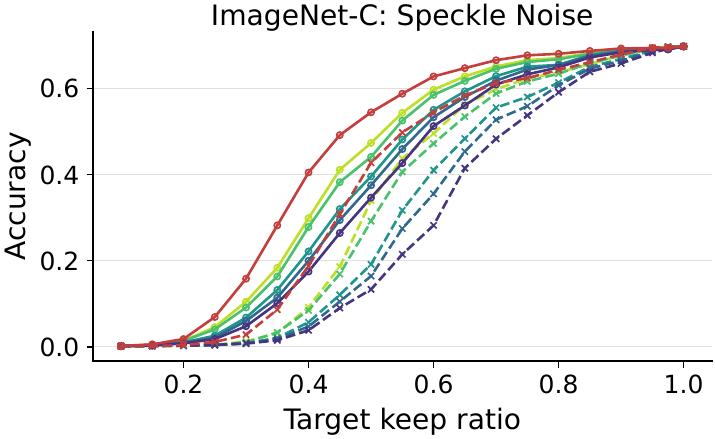}
        \label{fig:imagenetc_speckle_noise}
    \end{subfigure}\hfill
    \begin{subfigure}[t]{0.32\linewidth}
        \centering
        \includegraphics[width=\linewidth]{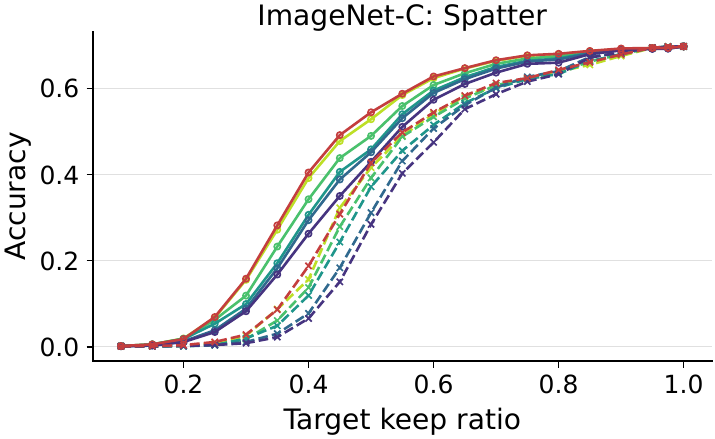}
        \label{fig:imagenetc_spatter}
    \end{subfigure}\hfill
    \begin{subfigure}[t]{0.32\linewidth}
        \centering
        \includegraphics[width=\linewidth]{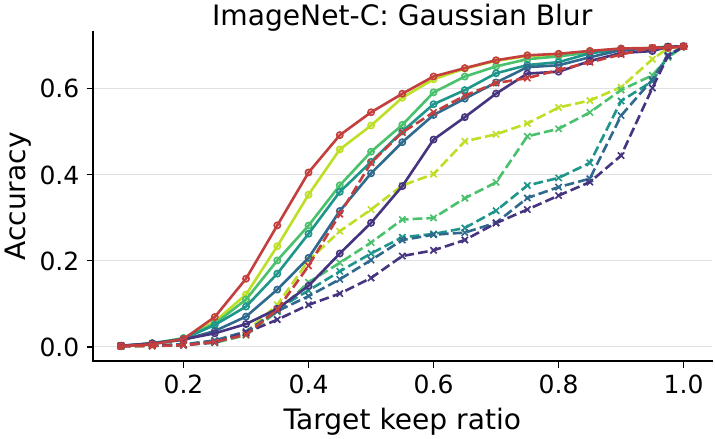}
        \label{fig:imagenetc_gaussian_blur}
    \end{subfigure}

    \vspace{0.8em}

    \begin{subfigure}[t]{0.32\linewidth}
        \centering
        \includegraphics[width=\linewidth]{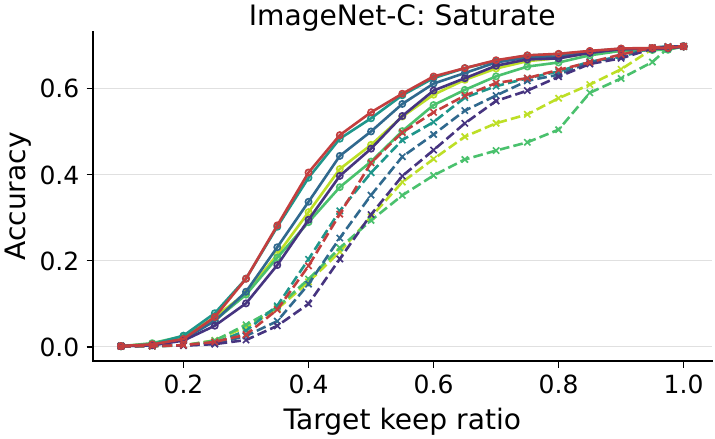}
        \label{fig:imagenetc_saturate}
    \end{subfigure}

    \vspace{0.2em}
    \includegraphics{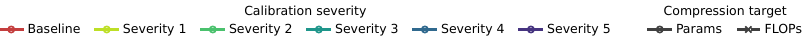}

    \caption{Compression--accuracy curves on ResNet-18 when calibration data are drawn from ImageNet-C and evaluation is performed on the standard ImageNet-1K validation set. Each panel corresponds to a ImageNet-C corruption type from the remaining weather, digital, and extra corruption categories. Notation as in \cref{fig:cifar10c_all_corruptions}.}
    \label{fig:imagenetc_all_corruptions_part2}
\end{figure}

\begin{figure}[ht]
    \centering

    \begin{subfigure}[t]{0.32\linewidth}
        \centering
        \includegraphics[width=\linewidth]{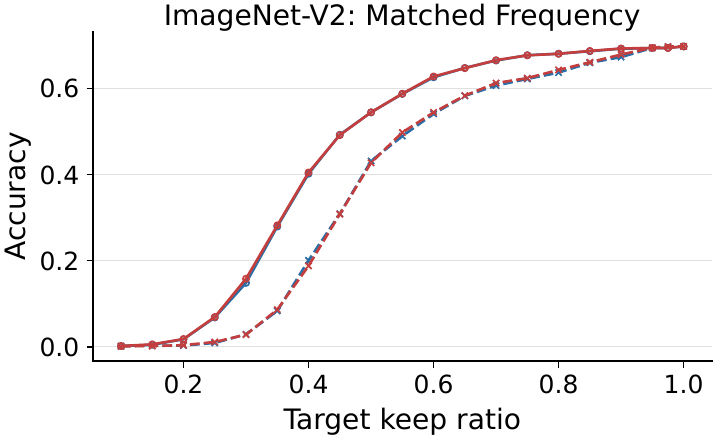}
        \label{fig:imagenetv2_matched_frequency}
    \end{subfigure}\hfill
    \begin{subfigure}[t]{0.32\linewidth}
        \centering
        \includegraphics[width=\linewidth]{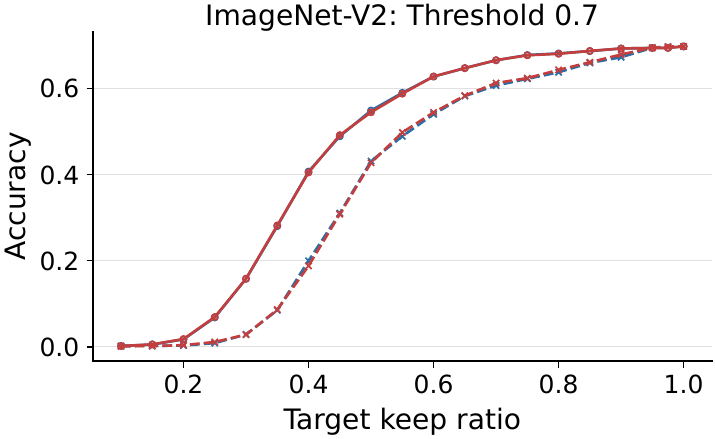}
        \label{fig:imagenetv2_threshold_07}
    \end{subfigure}\hfill
    \begin{subfigure}[t]{0.32\linewidth}
        \centering
        \includegraphics[width=\linewidth]{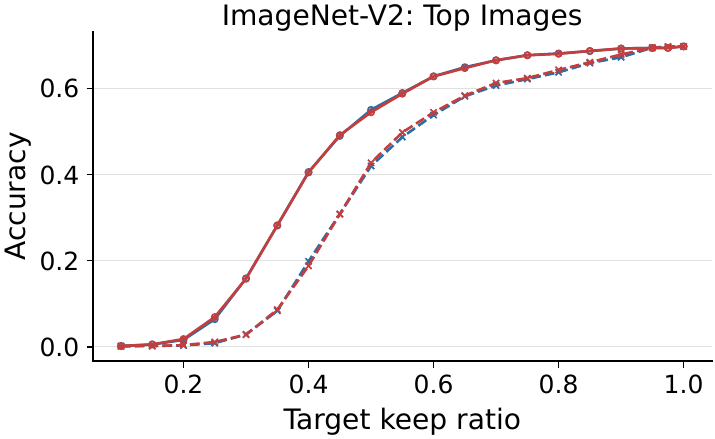}
        \label{fig:imagenetv2_topimages}
    \end{subfigure}

    \vspace{0.2em}
    \includegraphics{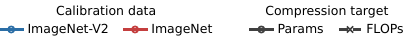}

    \caption{Compression--accuracy curves on ResNet-18 when calibration data are drawn from ImageNet-V2 and evaluation is performed on the standard ImageNet-1K validation set. Each panel corresponds to a ImageNet-V2 variant. Notation as in \cref{fig:cifar10c_all_corruptions}.}
    \label{fig:imagenetv2_variants}
\end{figure}

For CIFAR-10, we use ResNet-20 and draw calibration data from CIFAR-10-C \citep{hendrycks2018benchmarking}, while evaluating on the clean CIFAR-10 test set. For ImageNet-1K, we use ResNet-18 and draw calibration data from ImageNet-C \citep{hendrycks2018benchmarking} or ImageNet-V2 \citep{pmlr-v97-recht19a}, while evaluating on the standard ImageNet-1K validation set. Results for CIFAR-10 appear in \cref{fig:cifar10c_all_corruptions}, results for ImageNet-C appear in \cref{fig:imagenetc_all_corruptions_part1,fig:imagenetc_all_corruptions_part2}, and results for ImageNet-V2 appear in \cref{fig:imagenetv2_variants}.

For corruption-based datasets, the results vary substantially depending on the corruption type and severity. While some have minor effects on the effectiveness of BALF, others substantially degrade performance. However, performance on the ImageNet-V2 experiment suggests that BALF is robust under natural distribution shifts.

\subsection{Speedup of Low-Rank Layers in Practice}\label{subsec:practical_speedup}

This section discusses the relationship between our compression results (in terms of FLOPs) and the end-to-end speedup obtained in practice. We use basic PyTorch implementations of low-rank operators. In particular, we substitute each original layer with two sequential layers, as described in \cref{sec:thorough_decomposition}, without any specialized implementation.

We measure the throughput and latency of two ImageNet-1K models (ResNet-50 and ViT-B/16) on an RTX 2080 Ti and an A100 GPU at different compression ratios. Every batch normalization (BN) \citep{pmlr-v37-ioffe15} layer is first removed from the ResNet-50 model so as to mimic the usual inference scenario (where BN layers are folded). Our method does not interfere with BN folding.

\begin{figure}[ht]
    \centering

    \begin{subfigure}{0.32\textwidth}
        \centering
        \includegraphics[width=\linewidth]{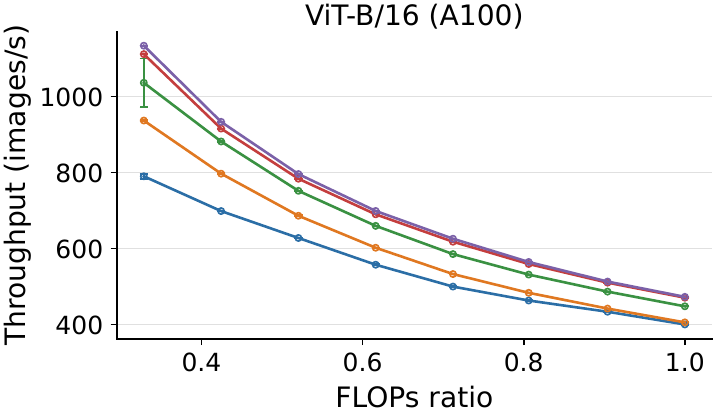}
        \label{fig:vit_throughput_a100}
    \end{subfigure}\hfill
    \begin{subfigure}{0.32\textwidth}
        \centering
        \includegraphics[width=\linewidth]{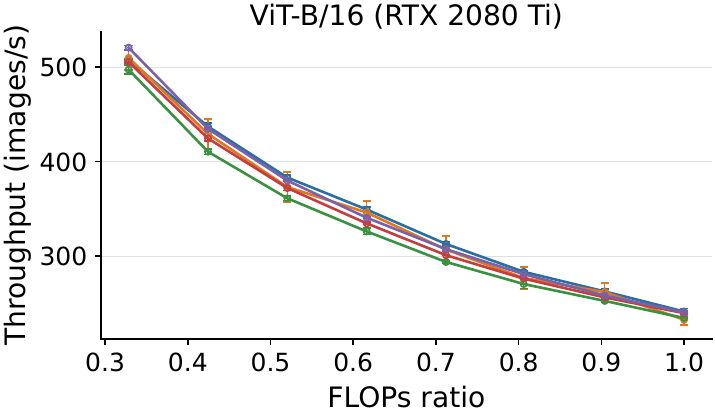}
        \label{fig:vit_throughput_rtx2080ti}
    \end{subfigure}\hfill
    \begin{subfigure}{0.32\textwidth}
        \centering
        \includegraphics[width=\linewidth]{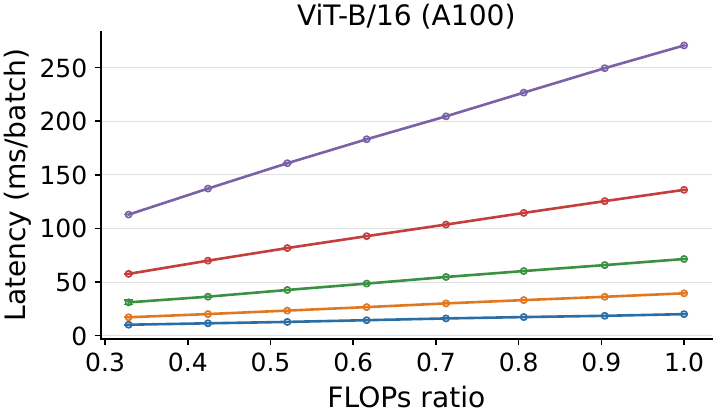}
        \label{fig:vit_latency_a100}
    \end{subfigure}

    \vspace{0.3em}

    \begin{subfigure}{0.32\textwidth}
        \centering
        \includegraphics[width=\linewidth]{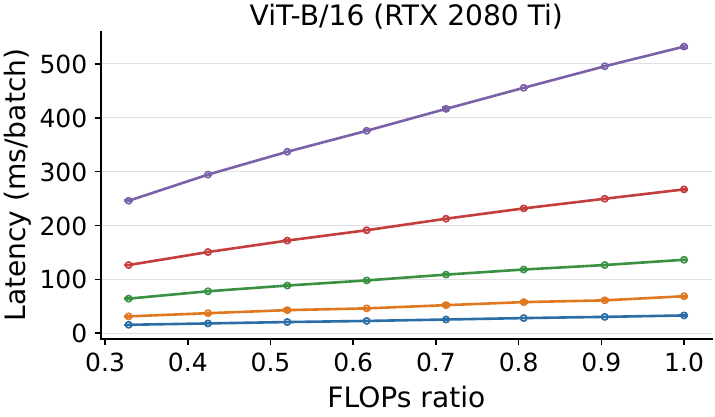}
        \label{fig:vit_latency_rtx2080ti}
    \end{subfigure}\hfill
    \begin{subfigure}{0.32\textwidth}
        \centering
        \includegraphics[width=\linewidth]{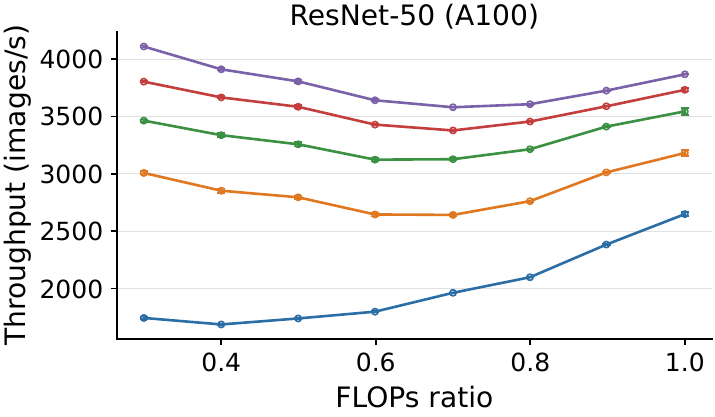}
        \label{fig:resnet50_throughput_a100}
    \end{subfigure}\hfill
    \begin{subfigure}{0.32\textwidth}
        \centering
        \includegraphics[width=\linewidth]{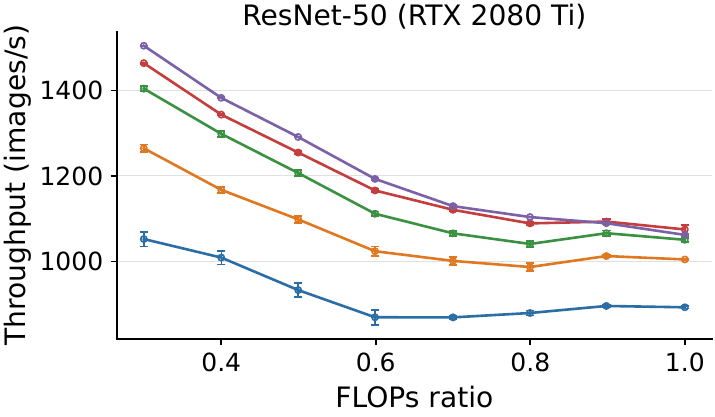}
        \label{fig:resnet50_throughput_rtx2080ti}
    \end{subfigure}

    \vspace{0.3em}

    \begin{subfigure}{0.32\textwidth}
        \centering
        \includegraphics[width=\linewidth]{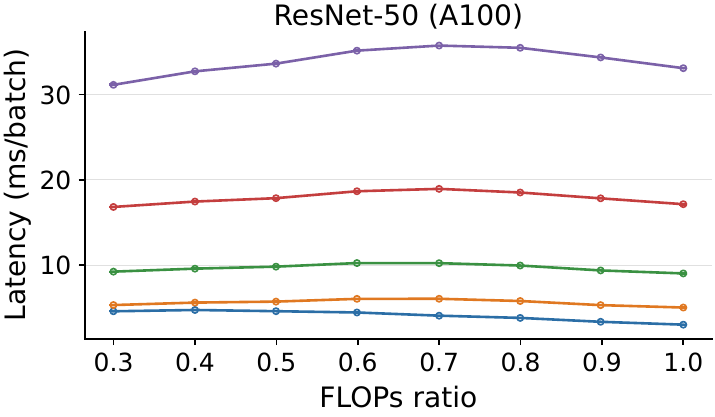}
        \label{fig:resnet50_latency_a100}
    \end{subfigure}
    \hspace{0.03\textwidth}
    \begin{subfigure}{0.32\textwidth}
        \centering
        \includegraphics[width=\linewidth]{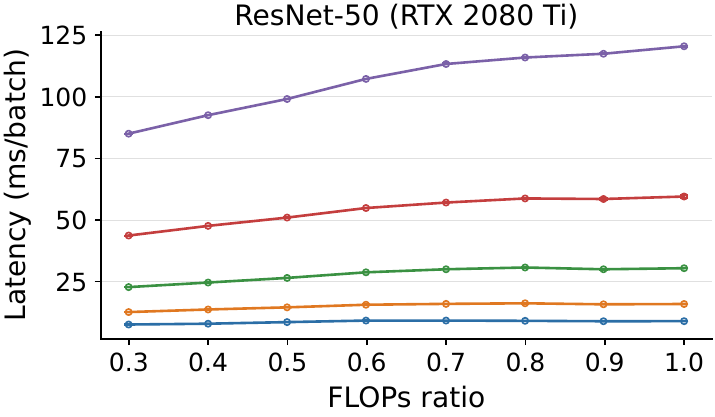}
        \label{fig:resnet50_latency_rtx2080ti}
    \end{subfigure}

    \vspace{0.2em}
    \includegraphics{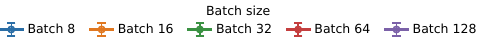}

    \caption{Throughput and latency of ViT-B/16 and ResNet-50 compressed with BALF, on NVIDIA A100 and RTX 2080 Ti GPUs across batch sizes. The FLOPs ratio denotes the fraction of the original model’s FLOPs retained after compression (e.g., 0.6 means 60\% retained).}
    \label{fig:throughput_latency}
\end{figure}

We find that compression results translate well to actual speedups (see \cref{fig:throughput_latency}) only in some circumstances. While matrix-multiplication-dominant models such as ViT obtain better speedups, convolutional layers fare worse than linear layers in this regard, and the theoretical gains are not fully exploited. In particular, on the A100 GPU, no consistent speedup is obtained for ResNet-50.

We expect that specialized operator implementations, including those found in software for edge devices, might be able to fully exploit the compression in terms of speedup. In fact, \citet{sui2024elrtefficientlowranktraining}, for instance, were able to obtain near-perfect practical speedups from FLOP reductions on ASICs and FPGAs on ResNet-50 with low-rank layers.

We note that this is, in general, also a persistent problem with other compression methods (see, e.g., \citet{narshana2023dfpc}, where, without specialized implementations, the actual speedups are significantly lower than the FLOPs reductions). We leave the exploration of more efficient implementations for future work.

\FloatBarrier
\section{Runtime and Memory Footprint of BALF}\label{sec:runtime_impl}

Our implementation of BALF is conceptually similar to \cref{alg:balf_overview}, but there are some implementation details that are important to discuss. We first provide a high-level overview of the implementation details:

\begin{enumerate}
\item First, given a calibration dataset, we gather the activation moments. The model runs on the GPU, but the activation moments are offloaded to CPU memory to reduce GPU memory usage and allow compression on lower-end GPUs. Once activation moments are gathered, they can be cached (on disk) for subsequent compression configurations.

\item Then, whitening matrices and factors are computed on the GPU. These are cached on disk to maintain low memory usage and to avoid recomputation in future runs.

\item Our rank allocator uses the singular values (computed in the preceding step) to decide how much to compress each layer.

\item Then, we iterate through the model and substitute the layers with their truncated versions.
\end{enumerate}

\renewcommand{\arraystretch}{1.1}
\begin{table}[ht]
\caption{Runtime breakdown and memory usage for different models. All timings are in seconds. Act. denotes activation moment collection, Fact.+Whit. denotes factorization and whitening, Replace denotes layer replacement, Misc denotes miscellaneous overhead, and Peak Mem. denotes peak CUDA memory usage during the run in GiB. We report mean and standard deviation across 3 runs.}
\label{table:timings}
\centering
\footnotesize
\begin{tabular}{lccccccc}
\hline
Model & Act. & Fact.+Whit. & Solver & Replace  & Misc & Total & Peak Mem. \\
\hline
ResNet-18 & $78.83_{\pm 0.41}$ & $5.74_{\pm 0.11}$ & $0.03_{\pm 0.01}$ & $0.72_{\pm 0.03}$ & $0.02_{\pm 0.00}$ & $85.35_{\pm 0.26}$ & $1.16_{\pm 0.00}$ \\
ResNet-50 & $97.65_{\pm 0.41}$ & $5.32_{\pm 0.28}$ & $0.06_{\pm 0.00}$ & $0.96_{\pm 0.01}$ & $0.00_{\pm 0.00}$ & $103.98_{\pm 0.48}$ & $1.34_{\pm 0.00}$ \\
MobileNetV2 & $41.54_{\pm 0.77}$ & $0.89_{\pm 0.09}$ & $0.05_{\pm 0.00}$ & $0.14_{\pm 0.00}$ & $0.00_{\pm 0.00}$ & $42.61_{\pm 0.68}$ & $2.23_{\pm 0.00}$ \\
ResNeXt-50 (32$\times$4d) & $56.97_{\pm 0.93}$ & $3.91_{\pm 0.06}$ & $0.06_{\pm 0.01}$ & $0.41_{\pm 0.01}$ & $0.00_{\pm 0.00}$ & $61.35_{\pm 0.98}$ & $2.29_{\pm 0.00}$ \\
ResNeXt-101 (32$\times$8d) & $181.22_{\pm 1.85}$ & $19.85_{\pm 0.44}$ & $0.13_{\pm 0.00}$ & $1.58_{\pm 0.08}$ & $0.01_{\pm 0.00}$ & $202.79_{\pm 1.92}$ & $4.45_{\pm 0.00}$ \\
ViT-B/16 & $112.14_{\pm 0.47}$ & $10.59_{\pm 0.19}$ & $0.11_{\pm 0.03}$ & $1.33_{\pm 0.12}$ & $0.00_{\pm 0.00}$ & $124.17_{\pm 0.57}$ & $0.83_{\pm 0.00}$ \\
DeiT-B/16 & $122.44_{\pm 6.79}$ & $9.87_{\pm 0.24}$ & $0.06_{\pm 0.00}$ & $1.20_{\pm 0.06}$ & $0.00_{\pm 0.00}$ & $133.57_{\pm 7.07}$ & $0.83_{\pm 0.00}$ \\
\hline
\end{tabular}
\end{table}
\renewcommand{\arraystretch}{1}
For a thorough examination of the implementation, we refer the reader to our code.\footnote{\url{\githuburl}} \Cref{table:timings} presents the runtime breakdown and peak memory usage of our method on different models on the ImageNet-1K dataset, evaluated on an RTX 2080 Ti GPU. We use 8192 calibration samples (as in our main experiments), divided into batches of 64 samples. All other parameters match those used in the main experiments. We note that the peak memory usage typically occurs during the computation of activation moments, and that using smaller batch sizes can further reduce it.

The only hyperparameters that are expected to affect the runtime of our method are the number of calibration samples and the number of solver iterations. For the first, it is only expected to affect (linearly) the activation-gathering phase (Act. in \cref{table:timings}).

Second, the number of solver iterations affects (linearly) the time it takes to produce a solution (Solver in \cref{table:timings}). Its runtime is negligible (less than 0.15 seconds in the runs of this section, using 300 iterations).

Finally, we reiterate that the most expensive steps (activation moment gathering and factorization/whitening) are cached, allowing one to obtain different models with different compression configurations in a matter of seconds on average, even on lower-end hardware, after the first run. We also note that our implementation was not particularly optimized (as compression time is not a bottleneck in our evaluations), and that further runtime gains are probably attainable.

\FloatBarrier
\section{A Post-Hoc Interpretation via \citet{shumaylov2025informationgeometryiterativeoptimization}}\label{sec:post_hoc_info_geometry}
Recent work by \citet{shumaylov2025informationgeometryiterativeoptimization} provides interesting insights into the factorization of neural networks from an information-geometric point of view. BALF has an interesting interpretation through the lens of their framework.

Let $\widehat{\tW}$ denote the low-rank approximation of a parameter tensor. They posit that general activation-aware factorization approaches solve an optimization problem equivalent to minimizing the squared output mismatch under an i.i.d. Gaussian noise model with identity covariance. Translating that to our framework, we obtain
\begin{align*}
\arg \min_{\{\widehat{\tW}{}^l\} \in \widehat{\mathcal{M}}} 
&\;\sum_{l=1}^L
  \left\|
    \ExprInp_l(\tX^{l-1}) \ExprParam_l(\tW^l) 
    - 
    \ExprInp_l(\tX^{l-1}) \ExprParam_l(\widehat{\tW}{}^l)
  \right\|_F^2
\\
= \arg \min_{\{\widehat{\tW}{}^l\} \in \widehat{\mathcal{M}}} 
&\;\sum_{l=1}^L \sum_{g=1}^{G_l} 
  \left\|
    \ExprInp_l(\tX^{l-1})_{g,:,:} \ExprParam_l(\tW^l)_{g,:,:}
    - 
    \ExprInp_l(\tX^{l-1})_{g,:,:} \ExprParam_l(\widehat{\tW}{}^l)_{g, :, :}
  \right\|_F^2
\\
= \arg \min_{\{\widehat{\tW}{}^l\} \in \widehat{\mathcal{M}}}  
&\;\sum_{l=1}^L \sum_{g=1}^{G_l} \sum_{n=1}^{N_l}
  \left\|
    \ExprInp_l(\tX^{l-1})_{g,n,:} \ExprParam_l(\tW^l)_{g,:,:}
    - 
    \ExprInp_l(\tX^{l-1})_{g,n,:} \ExprParam_l(\widehat{\tW}{}^l)_{g, :, :}
  \right\|_2^2
\\
= \arg \min_{\{\widehat{\tW}{}^l\} \in \widehat{\mathcal{M}}}  
&\;D_\mathrm{{KL}}\!\bigl(
    p_{\tW}\,\|\,p_{\smash{\widehat{\tW}}\vphantom{\tW}}
  \bigr),
\end{align*}
with
\[
p_{\tW}(\cdot)
=
\prod_{l=1}^L\prod_{g=1}^{G_l} \prod_{n=1}^{N_l}
\mathcal{N}\!\left(
\cdot \,\middle|\,
\ExprInp_l(\tX^{l-1})_{g,n} \ExprParam_l(\tW^{l})_g,
\mI
\right),
\]
and
\[
p_{\smash{\widehat{\tW}}\vphantom{\tW}}(\cdot)
=
\prod_{l=1}^L\prod_{g=1}^{G_l} \prod_{n=1}^{N_l}
\mathcal{N}\!\left(
\cdot \,\middle|\,
\ExprInp_l(\tX^{l-1})_{g,n} \ExprParam_l(\widehat{\tW}{}^{l})_g,
\mI
\right).
\]
In the preceding expressions, the $n$ index corresponds to the outer dimension (e.g., with $B H_o W_o = N$ for convolutional layers), $g$ to the groups, and $l$ to the layers. The optimization problem generally assumes ranks are fixed ($\widehat{\mathcal{M}}$ denotes a model with layers truncated to fixed ranks, assumed to be given in advance), and it is only concerned with finding the best low-rank projections.

On the other hand, the problem BALF solves can be formulated as
\begin{align*}
\arg \min_{\{\widehat{\tW}{}^{l}\}  \in \widehat{\mathcal{M}}: C(\widehat{\mathcal{M}}) \leq C_{\max}}
&\;
\sum_{l=1}^L
\frac{
  \left\|
    \ExprInp_l(\tX^{l-1}) \ExprParam_l(\tW^{l})
    - \ExprInp_l(\tX^{l-1}) \ExprParam_l(\widehat{\tW}{}^{l})
  \right\|_{F}^{2}
}{
  \left\|
    \ExprInp_l(\tX^{l-1}) \ExprParam_l({\tW}^{l})
  \right\|_{F}^{2}
}
\\
= \arg \min_{\{\widehat{\tW}{}^{l}\}  \in \widehat{\mathcal{M}}: C(\widehat{\mathcal{M}}) \leq C_{\max}}
&\;
\sum_{l=1}^L
\sum_{g=1}^{G_l}
\frac{
  \left\|
    \ExprInp_l(\tX^{l-1})_{g,:,:} \ExprParam_l(\tW^{l})_{g,:,:}
    - \ExprInp_l(\tX^{l-1})_{g,:,:} \ExprParam_l(\widehat{\tW}{}^{l})_{g,:,:}
  \right\|_{F}^{2}
}{
  \left\|
    \ExprInp_l(\tX^{l-1}) \ExprParam_l(\tW^{l})
  \right\|_{F}^{2}
}
\\
= \arg \min_{\{\widehat{\tW}{}^{l}\}  \in \widehat{\mathcal{M}}: C(\widehat{\mathcal{M}}) \leq C_{\max}}
&\;
\sum_{l=1}^L
\sum_{g=1}^{G_l} \sum_{n=1}^{N_l}
\frac{
  \left\|
    \ExprInp_l(\tX^{l-1})_{g,n,:} \ExprParam_l(\tW^{l})_{g,:,:}
    - \ExprInp_l(\tX^{l-1})_{g,n,:} \ExprParam_l(\widehat{\tW}{}^{l})_{g,:,:}
  \right\|_{2}^{2}
}{
  \left\|
    \ExprInp_l(\tX^{l-1}) \ExprParam_l({\tW}^{l})
  \right\|_{F}^{2}
}
\\
= \arg \min_{\{\widehat{\tW}{}^l\}  \in \widehat{\mathcal{M}}: C(\widehat{\mathcal{M}}) \leq C_{\max}}
&\;D_\mathrm{{KL}}\!\bigl(
    p_{\tW}\,\|\,p_{\smash{\widehat{\tW}}\vphantom{\tW}}
  \bigr),
\end{align*}

with 
\[
p_{\tW}(\cdot)
=
\prod_{l=1}^L\prod_{g=1}^{G_l}\prod_{n=1}^{N_l}
\mathcal{N}\!\left(
\cdot \,\middle|\,
\ExprInp_l(\tX^{l-1})_{g,n} \ExprParam_l(\tW^l)_g,
\left\|
\big(\ExprInp_l(\tX^{l-1}) \ExprParam_l(\tW^{l})\big)
\right\|_{F}^{2}\,\mI
\right),
\]
and
\[
p_{\smash{\widehat{\tW}}\vphantom{\tW}}(\cdot)
=
\prod_{l=1}^L\prod_{g=1}^{G_l}\prod_{n=1}^{N_l}
\mathcal{N}\!\left(
\cdot \,\middle|\,
\ExprInp_l(\tX^{l-1})_{g,n} \ExprParam_l(\widehat{\tW}{}^{l})_g,
\left\|
\big(\ExprInp_l(\tX^{l-1}) \ExprParam_l(\tW^{l})\big)
\right\|_{F}^{2}\,\mI
\right).
\]

There are two main differences. First, BALF also works under the Gaussian assumption, but with the additional nuance that the variances depend on the per-layer output magnitude (in terms of the squared Frobenius norm). Second, the optimization problem is not only concerned with the low-rank projection scheme given a fixed rank per layer; the selection of ranks under some complexity budget is, itself, baked into the optimization problem. (Note that, strictly speaking, BALF solves the problem approximately in practice.)

We believe this point of view might be an interesting avenue for future research.

\section{Deferred Proofs}\label{sec:appendix_proofs}
In this section, we provide full proofs for the results in the main text that were not covered in other sections of the appendix. Results are restated for the convenience of the reader.

\subsection{Proof of \cref{prop:general_activation_aware_distortion}}
\GeneralActivationAwareDistortionProp*
\begin{proof}

First, observe that 
\begin{equation}\label{eq:observation}
\|\mathsf{cat}(\mA_1,\ldots,\mA_K)\|_F^2
= \sum_{i=1}^K \|\mA_i\|_F^2,
\end{equation}
where $\mathsf{cat}$ denotes the operation that yields a third-order tensor from a batch of matrices (of the same shape). Therefore, for a batch of distortion matrices, we may analyze each term independently and sum their contributions. Moreover,
\begin{align*}
\bigl\| f(\tX;\tW) - f\bigl(\tX;\AATrunc{P}{\tW}\bigr) \bigr\|_F
&= \bigl\| \ExprOut\!\bigl(\ExprInp(\tX)\,\ExprParam(\tW)\bigr)
      - \ExprOut\!\bigl(\ExprInp(\tX)\,\AATrunc{P}{\ExprParam(\tW)}\bigr) \bigr\|_F \\
&= \bigl\| \ExprInp(\tX)\,\ExprParam(\tW)
      - \ExprInp(\tX)\,\AATrunc{P}{\ExprParam(\tW)} \bigr\|_F,
\end{align*}
since $\ExprOut$ is a composition of reshapes and permutes, which preserve the Frobenius norm.

Fix a group index $g$ (omitted from the notation below for simplicity). For reference, the dimensions of the different elements are: $\ExprInp(\tX) \in \Rspace{N,I}$, $\mM \in \Rspace{I,I}$, $\mM^{+} \in \Rspace{I,I}$, and $\ExprParam(\tW) \in \Rspace{I,O}$.

When using activation-aware factorization and truncating to rank $P$, it follows that
\begin{align*}
\ell_{(g)}^{\mathrm{out}} &= \frac{1}{B} \left\| \ExprInp(\tX)\ExprParam(\tW) - \ExprInp(\tX) \mM \SVDTrunc{P}{\mM^{+}\ExprParam(\tW)}\right\|^2_F \notag \\
    &\mathrel{\underset{\text{(a)}}{=}} \frac{1}{B}\left\| \ExprInp(\tX)\mM \left( \mM^{+} \ExprParam(\tW) - \SVDTrunc{P}{\mM^{+}\ExprParam(\tW)} \right) \right\|_F^2 \\
    &= \frac{1}{B}\left\| \ExprInp(\tX)\mM  \mU\widehat{\mSigma} \mV^\top \right\|^2_F \\
    &= \frac{1}{B}\operatorname{tr}\left(  \mV\widehat{\mSigma}^\top \mU^\top \mM^\top \ExprInp(\tX)^\top \ExprInp(\tX) \mM  \mU\widehat{\mSigma} \mV^\top\right) \\
    &\mathrel{\underset{\text{(b)}}{=}} \frac{N}{B}\operatorname{tr}\left( \mV\widehat{\mSigma}^\top \widehat{\mSigma} \mV^\top\right) \\
    &= \frac{N}{B}\sum_{i=P+1}^U\bm{\sigma}_i^2,
\end{align*}
where the SVD is taken of $\mM^{+}\ExprParam(\tW)$, and $\widehat{\mSigma}$ corresponds to the tail of $\mSigma$ after truncating the singular values (i.e., it is equal to $\mSigma$ but with the first $P$ diagonal elements set to 0).

For step (a), note that $\ExprInp(\tX) \mM \mM^+ = \ExprInp(\tX)$ (see \cref{sec:app_formal_uwm}).

For step (b), note the following. First, recall from the definition of $\mM$ (see \cref{sec:app_formal_uwm}) that $\mM^\top \ExprInp(\tX)^\top \ExprInp(\tX) \mM = N\overline{\mI}_R$ (where $R$ is the rank of $\ExprInp(\tX)$); and that $\mM$ has its last $I - R$ columns set to 0, meaning $\mM^+$ has its last $I - R$ rows set to 0.

As $\mU\widehat{\mSigma}\mV^\top$ corresponds to the SVD tail of $\mM^{+}\ExprParam(\tW) \in \Rspace{I, O}$, and $\mM^{+}\ExprParam(\tW)$ has its last $I - R$ rows set to 0, all left singular vectors associated with nonzero singular values lie in the range of $\overline{\mI}_R$. Hence, $\overline{\mI}_R\mU\widehat{\mSigma}=\mU\widehat{\mSigma}$, or equivalently $\widehat{\mSigma}^\top\mU^\top\overline{\mI}_R\mU\widehat{\mSigma}=\widehat{\mSigma}^\top\widehat{\mSigma}$. This results in $\widehat{\mSigma}^\top\mU^\top \mM^\top \ExprInp(\tX)^\top \ExprInp(\tX) \mM \mU\widehat{\mSigma} = N\widehat{\mSigma}^\top\mU^\top\overline{\mI}_R\mU\widehat{\mSigma} = N\widehat{\mSigma}^\top\widehat{\mSigma}$.

From the observation in \cref{eq:observation}, it follows that the total distortion is the sum across each group's distortion
\[
\ell^{\mathrm{out}}(\tW, \AATrunc{P}{\tW}) = \sum_{g = 1}^G \ell_{(g)}^{\mathrm{out}} = \frac{N}{B}\sum_{g=1}^{G}{\sum_{i=P+1}^U\bm{\sigma}_{g,i}^2},
\]
which completes the proof.

\end{proof}

\subsection{Proof of \cref{prop:optimality}}

\begin{lemma}\label{lemma:optimality_prop_lemma}
Let $\mY \in \Rspace{M, N}$ satisfy $\mY^\top \mY = \alpha \mI_N$ with $\alpha>0$. Then, for any $\mX \in \Rspace{N, K}$,
\[
\bm{\sigma}_i(\mY \mX) = \sqrt{\alpha}\bm{\sigma}_i(\mX)
\]
for all $i$, with singular values padded by zeros if necessary. Consequently, for any $P \ge 1$,
\[
\sum_{i=1}^P \bm{\sigma}_i^2(\mY \mX) = \alpha\sum_{i=1}^P \bm{\sigma}_i^2(\mX).
\]
\end{lemma}
\begin{proof}
We have
\[
(\mY \mX)^\top(\mY \mX) = \mX^\top(\mY^\top \mY)\mX = \alpha\mX^\top\mX.
\]
Hence, the eigenvalues of $(\mY \mX)^\top(\mY \mX)$ are exactly $\alpha$ times those of $\mX^\top\mX$, including zeros. Therefore, $\bm{\sigma}_i(\mY \mX)=\sqrt{\alpha}\bm{\sigma}_i(\mX)$ for all $i$. Squaring and summing the first $P$ singular values gives the claim.
\end{proof}

\OptimalityProp*
\begin{proof}
It suffices to prove the result for a single group. Fix a group (its index is omitted from the notation for brevity), and let $\mW = \ExprParam(\tW) \in \Rspace{I, O}, \mX=\ExprInp(\tX)\in\Rspace{N,I}$, $\mM\in\Rspace{I,I}$, and $\mM^+\in\Rspace{I,I}$ denote the items corresponding to said group, and denote the rank of $\mX$ by $R$.

We recall the following facts; see \cref{sec:app_formal_uwm}. First, $\mM$ has its last $I-R$ columns set to $0$, which forces $\mM^+$ to have its last $I-R$ rows set to $0$. Hence, $\mM \mM^+ = \mM\mI_{:,:R}\mI_{:R,:}\mM^+$. Moreover, from the definition of $\mM$, we have $\mM^\top \mX^\top \mX \mM = N \overline{\mI}_R$, and therefore, it holds that $\mI_{:R,:}\mM^\top \mX^\top \mX \mM\mI_{:,:R} = N \mI_R$. It is also true that $\mX\mM\mM^+ = \mX$.

It follows that
\[
\sum_{i=1}^P \bm{\sigma}^2_i(\mX \mW) = \sum_{i=1}^P \bm{\sigma}^2_i(\mX \mM  \mM^+ \mW) = \sum_{i=1}^P \bm{\sigma}^2_i(\mX \mM\mI_{:,:R}\mI_{:R,:}\mM^+ \mW) = N \sum_{i=1}^P \bm{\sigma}^2_i(\mM^+ \mW),
\]
where the last equality follows from \cref{lemma:optimality_prop_lemma}.

Hence,
\begin{align*}
\|\ExprInp(\tX)\ExprParam(\tW)-\SVDTrunc{P}{\ExprInp(\tX)\ExprParam(\tW)}\|^2_F
  &= \sum_{i=P+1}^U \bm{\sigma}_i^2(\mX \mW) \\
  &= N \sum_{i=P+1}^U \bm{\sigma}_i^2(\mM^+ \mW) \\
  &= \|\ExprInp(\tX)\ExprParam(\tW)-\ExprInp(\tX)\AATrunc{P}{\ExprParam(\tW)}\|^2_F,
\end{align*}
where the last equality follows from \cref{prop:general_activation_aware_distortion}.

Taking square roots completes the proof.

\end{proof}

\end{document}